\documentclass[10pt,twocolumn,letterpaper]{article}

\usepackage{cvpr}              % To produce the CAMERA-READY version
\definecolor{cvprblue}{rgb}{0.21,0.49,0.74}
\usepackage[pagebackref,breaklinks,colorlinks,allcolors=cvprblue]{hyperref}

\def\paperID{15118} % *** Enter the Paper ID here
\def\confName{CVPR}
\def\confYear{2025}

\usepackage[accsupp]{axessibility}
\usepackage[ruled,vlined,linesnumbered]{algorithm2e}
\usepackage{amsmath}
\usepackage{amsthm}
\usepackage{booktabs}
\usepackage{multirow}
\usepackage{subcaption}
\usepackage{graphicx}
\usepackage{dsfont}
\newcommand{\method}{T-CIL}
\DeclareMathOperator*{\argmax}{arg\,max}
\DeclareMathOperator*{\argmin}{arg\,min}

\title{\method: Temperature Scaling using Adversarial Perturbation for Calibration in Class-Incremental Learning}

\author{Seong-Hyeon Hwang\quad Minsu Kim\quad Steven Euijong Whang\thanks{Corresponding Author}\\
KAIST\\
{\tt\small \{sh.hwang, ms716, swhang\}@kaist.ac.kr}
}
\begin{document}
\maketitle

\begin{abstract}
We study model confidence calibration in class-incremental learning, where models learn from sequential tasks with different class sets. While existing works primarily focus on accuracy, maintaining calibrated confidence has been largely overlooked. Unfortunately, most post-hoc calibration techniques are not designed to work with the limited memories of old-task data typical in class-incremental learning, as retaining a sufficient validation set would be impractical. Thus, we propose \method{}, a novel temperature scaling approach for class-incremental learning without a validation set for old tasks, that leverages adversarially perturbed exemplars from memory. Directly using exemplars is inadequate for temperature optimization, since they are already used for training. The key idea of \method{} is to perturb exemplars more strongly for old tasks than for the new task by adjusting the perturbation direction based on feature distance, with the single magnitude determined using the new-task validation set. This strategy makes the perturbation magnitude computed from the new task also applicable to old tasks, leveraging the tendency that the accuracy of old tasks is lower than that of the new task. We empirically show that \method{} significantly outperforms various baselines in terms of calibration on real datasets and can be integrated with existing class-incremental learning techniques with minimal impact on accuracy.
\end{abstract}

\maketitle

\section{Introduction}
Nowadays it is essential for deep neural networks to deliver not only precise predictions, but also trustworthy confidence levels indicating the likelihood that predictions are correct. However, with the enhancement of the model's capabilities, there is a tendency for the model to exhibit excessive confidence in its predictions compared to the actual accuracy\,\citep{guo2017calibration}. This discrepancy between the confidence levels of the model and the actual accuracy makes model predictions unreliable to use for decision making. The discrepancy also results in severe failures in real-world safety-critical applications that require reliable models such as autonomous driving\,\citep{grigorescu2020survey} and medical diagnostics\,\citep{jiang2012calibrating}.

With the growing need for reliable models, confidence calibration \citep{guo2017calibration}, which adjusts confidence levels of the model to match their actual accuracies, has gained increasing attention. To enhance the calibration of deep neural networks, both post-hoc and training-time methods have been introduced. Post-hoc methods adjust the confidence levels of a model after training by applying a calibration function to the predictions made on a hold-out validation set, such as temperature scaling\,\citep{guo2017calibration} or histogram binning\,\citep{zadrozny2001obtaining}. Training-time methods refine model calibration during the training phase by altering the loss function\,\citep{mukhoti2020calibrating}, implementing label smoothing\,\citep{szegedy2016rethinking}, or using Mixup\,\citep{DBLP:conf/iclr/ZhangCDL18} training on interpolated data samples.

\begin{figure*}[t]
\centering
\includegraphics[width=\textwidth]{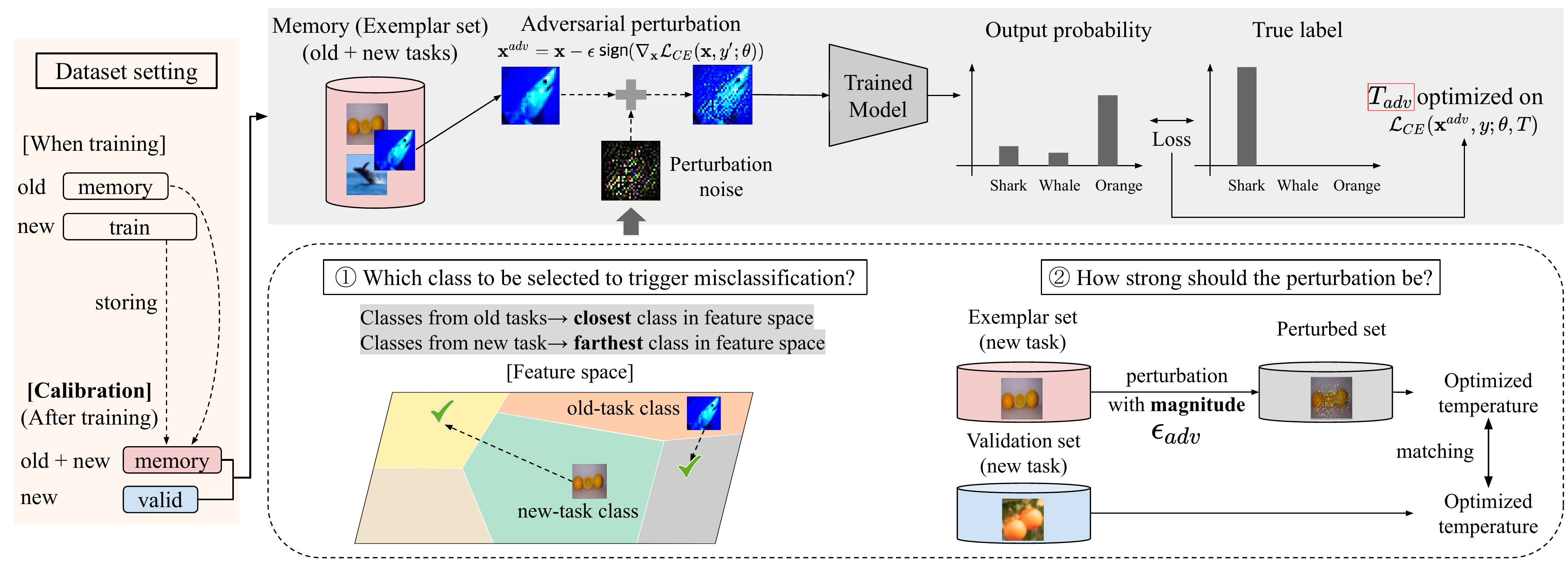}
\vskip -0.08in
\caption{Overview of our proposed framework, \method{}, a post-hoc calibration framework for class-incremental learning when a validation set only from the new task is available with memory. Our method leverages exemplars from memory by applying adversarial perturbations, whose direction and magnitude are determined based on feature distance and new task validation data, respectively. The temperature is then optimized using these perturbed exemplars.}
\label{fig:framework}
\vskip -0.1in
\end{figure*}

However, conventional post-hoc calibration approaches struggle in class-incremental learning scenarios, where models must continuously adapt to new classes with sufficient data while retaining performance on previously learned classes. Existing works for class-incremental learning\,\citep{rebuffi2017icarl, lopez2017gradient, castro2018end, wu2019large, yan2021dynamically, zhou2023deep} focus on preventing catastrophic forgetting\,\citep{mccloskey1989catastrophic}, losing knowledge of old tasks as they learn new ones. They often assume a limited memory size for old classes, as storing all data may be impractical due to memory constraints and privacy concerns. For post-hoc approaches, the primary challenge stems from this memory limitation. This limitation restricts the retention of a sufficient validation set from old tasks. 
If a portion of the stored data is reserved solely for validation and excluded from training, model accuracy on old tasks will likely decline due to the reduced amount of training data.

To effectively calibrate models in class-incremental learning, we propose \method{}, a novel temperature scaling approach that operates without validation data from old tasks (see Figure~\ref{fig:framework}). While exemplars from memory are the only available source of information for old tasks, directly using them for temperature optimization yields abnormally small values due to their use in training. Thus, we optimize the temperature on adversarially perturbed exemplars. We introduce adversarial perturbation of exemplars, designed with two key components: \textit{direction} and \textit{magnitude}.

The key idea of \method{} is to perturb exemplars more strongly for old tasks than for the new task by adjusting the perturbation direction based on feature distance, with the magnitude determined using the new-task validation set. While the perturbation levels can be controlled simply with two magnitude parameters for old and new tasks, determining the proper magnitude for old tasks is unrealistic without a validation set, as it is impossible to know how much the magnitude affects temperature optimization exactly. Therefore, we use a single magnitude determined using a new-task validation set and adjust the levels by setting the perturbation direction using target class selection. Since old tasks generally exhibit lower test accuracy than new tasks, we induce easier mispredictions on exemplars from old tasks than new task with the same magnitude. \method{} is guiding the model to misclassify old-task data into their closest classes in feature space, while mispredicting new-task data into the farthest classes. This strategy makes the perturbation magnitude computed from the new task also applicable to old tasks, leveraging the tendency that the accuracy of old tasks is lower than that of the new task.

We perform comprehensive experiments on various image classification datasets in class-incremental learning settings. We demonstrate the effectiveness of \method{} in terms of confidence calibration while minimally changing accuracy without a validation data from old tasks. We show that \method{} consistently outperforms existing post-hoc calibration methods and can be integrated with any class-incremental learning technique. Even if there is enough data to construct a validation set, this may result in a tradeoff in accuracy, which \method{} does not have.  For instance, when using Experience Replay (ER)\,\citep{chaudhry2019tiny} with 100\% of the memory for training exemplars, instead of reserving 25\% for validation, we observe an improvement in average accuracy on CIFAR-100 from 53.56\% to 56.25\%. This performance gain is comparable to the accuracy improvement achieved by EEIL\,\citep{castro2018end} (a method specifically designed to mitigate forgetting) over ER, but with the significant advantage of not requiring an additional old-task validation set. See Section~\ref{sec:compatibility} for more details.

\textbf{Summary of Contributions:} (1) We propose a new post-hoc calibration method, called \method{}, specifically designed for class-incremental learning; (2) We introduce a novel temperature optimization approach using adversarially perturbed exemplars without a validation set from old tasks; (3) We demonstrate the effectiveness of \method{} through comprehensive experiments, showing superior calibration performance and compatibility across various class-incremental learning methods.

\section{Related Work}
\paragraph{Confidence Calibration}
The purpose of confidence calibration is to narrow the discrepancy between a model's accuracy and its confidence levels. Existing calibration works can be categorized into two main types: training-time methods and post-hoc methods. Training-time methods consider calibration during model training. Using other types of loss functions instead of a vanilla cross-entropy loss are considered training-time methods\,\citep{mukhoti2020calibrating, ghosh2022adafocal}. Implicit regularizations are also widely used during training including label smoothing\,\citep{szegedy2016rethinking, liu2022devil, park2023acls} and Mixup\,\citep{DBLP:conf/iclr/ZhangCDL18, noh2023rankmixup}.

In comparison, our focus is on post-hoc approaches where the model's output logits are adjusted after training while preserving the order of predicted classes. Post-hoc methods include Platt scaling\,\citep{platt1999probabilistic}, histogram binning\,\citep{zadrozny2001obtaining}, dirichlet scaling\,\citep{kull2019beyond}, isotonic regression\,\citep{zadrozny2002transforming}. In particular, temperature scaling\,\citep{guo2017calibration} is a technique that scales the model's output logits using a temperature parameter determined on a validation set and is known to be effective in in-distribution contexts. Many variants\,\citep{zhang2020mix, ding2021local, joy2023sample} of temperature scaling have also been proposed. However, the performance of temperature scaling deteriorates in out-of-distribution scenarios. Despite this challenge, temperature scaling has been utilized in various settings including distribution shifts\,\citep{ovadia2019can}, distribution overlaps\,\citep{chidambaram2024limitations}, multi-domain\,\citep{yu2022robust}, out-of-domain\,\citep{choi2024consistency}, and domain drift situations\,\citep{tomani2021post}. More recently, \citep{li2024calibration} is the first calibration study in continual learning that uses some of the above existing methods along with an old-task validation set. All of these works assume access to the information of all classes via a validation set. In comparison, we assume that the validation set only contains data for classes from the new task and that any data from old tasks is unavailable. Under this condition, \method{} is the first effective temperature scaling without a validation set from old tasks.

\paragraph{Class-Incremental Learning (CIL)}
Class-incremental learning\,\citep{cauwenberghs2000incremental, zhou2002ensembling, Kuzborskij_2013_CVPR, zhou2023deep} is a type of continual learning in which new tasks are learned sequentially, each with unique sets of classes. The main objective in class-incremental learning research is to mitigate the catastrophic forgetting problem to maintain accuracy across tasks. From a data-centric perspective, some studies use a small number of exemplars from old tasks\,\citep{chaudhry2019tiny, aljundi2019online} to adjust the directions of the gradient\,\citep{lopez2017gradient, chaudhry2018efficient} or apply knowledge distillation\,\citep{rebuffi2017icarl, castro2018end, wu2019large, zhao2020maintaining}. For an architecture perspective, dynamic architecture expansion approaches\,\citep{yan2021dynamically} have been proposed. A particular study\,\citep{kang2020confidence} addresses model confidence within class-incremental learning, focusing mainly on the confidence of the latest task rather than the older ones. In contrast, our study focuses on recalibrating the confidence of both old and new tasks. While existing class-incremental learning approaches often overlook confidence calibration, \method{} specifically addresses this crucial aspect.

\section{Preliminaries}
\paragraph{Notation for CIL}
Suppose we are at the $t$-th incremental task in an offline class-incremental learning setting. Let $C_{t, old}=\sum_{j=1}^{t-1} C_j$ denote the number of classes from old tasks, where $C_j$ is the number of classes at $j$-th task, and $C_t$ represents the number of classes in the new task. We train a model to classify an image into a total of $C_{t, old} + C_t$ classes, combining both old and new tasks.

We consider a scenario after the model training at the $t$-th incremental task, as our focus is on the post-hoc process. Before training, we have a dataset of the new task, $\mathcal{D}_t = \{(\mathbf{x}_i, y_i)\}$, where $\mathbf{x}_i \in \mathcal{X}_t$ is an input sample and $y_i \in \mathcal{Y}_t = \{C_{t, old}+1, ..., C_{t, old}+C_t\}$ is its corresponding label. Also we obtain a memory set from the $t-1$-th task, $\mathcal{M}_{t-1} = \{(\mathbf{x}_j, y_j)\}$, containing a subset of data (exemplars) from old tasks, where $y_j \in \{1, ..., C_{t, old}\}$. We split $\mathcal{D}_t$ into a training set, $\mathcal{D}_{t, train}$, and a small validation set $\mathcal{D}_{t, valid}$. The memory is significantly smaller than the training set for the new task, such that $|\mathcal{D}_{t, train}|/C_t \gg |\mathcal{M}_{t-1}|/C_{t, old}$.
When training the $t$-th task, we use both $\mathcal{D}_{t,train}$ and $\mathcal{M}_{t-1}$. After training, we update the memory, denoted as $\mathcal{M}_t$, by storing a portion of $\mathcal{D}_{t, train}$ in it. Calibration is then performed using $\mathcal{D}_{t, valid}$ and $\mathcal{M}_{t}$, which contains data from both old and new tasks.

\paragraph{Notation for Calibration}
Let $f_\theta$ be a classifier with parameters $\theta$. $f_\theta$ outputs a logit $\mathbf{z}_i = f_\theta(\mathbf{x}_i)$, where $f_\theta : \mathcal{X} \rightarrow \mathbb{R}^K$ with dimension $K = C_{t, old} + C_t$. Let $\mathcal{L}$ be the loss function that measures how well the model's predictions $f_\theta(\mathbf{x}_i)$ matches the true label $y_i$. Especially, $\mathcal{L}_{CE}$ is the cross-entropy loss on a sample $(\mathbf{x}_i, y_i)$ defined as:
\begin{align}
\mathcal{L}_{CE}(\mathbf{x}_i, y_i ; \theta) = -\sum_{k=1}^K y_{i, k} \log p_{i, k}
\end{align}
where $p_{i, k} = \frac{e^{z_{i,k}}}{\sum_{j=1}^K e^{z_{i,j}}}$ is the output probability of the $k$-th class for the $i$-th input calculated by applying the softmax function to the logit $\mathbf{z}_i$. $\hat{p}_i$ = $\max_{k} p_{i, k}$ is the confidence level, and $\hat{y}_i = \argmax_{k} p_{i, k}$ is the predicted class of $\mathbf{x}_i$.

To measure how well a model is calibrated, we introduce the Expected Calibration Error (ECE)\,\citep{naeini2015obtaining} defined as $\mathbb{E}_{(\mathbf{x}_i, y_i) \sim \mathcal{P}}$ $[|P(\hat{y}_i = y_i|\hat{p}_i) - \hat{p}_i|]$. In practice, with a finite test set, we divide the interval [0, 1] into $B$ equal-with bins, where $B_i$ is the $i$-th bin representing the interval $\left(\frac{i-1}{B}, \frac{i}{B}\right]$. We compute the average accuracy, $acc(B_i) = \frac{1}{|B_i|} \sum_{j \in B_i} \mathds{1}(\hat{y}_j = y_j)$, where $\mathds{1}(\cdot)$ is an indicator function, and the average confidence, $conf(B_i) = \frac{1}{|B_i|} \sum_{j \in B_i} \hat{p}_j$. Then, ECE is calculated as $\sum^B_{i=1} \frac{|B_i|}{N} |acc(B_i) - conf(B_i)|$, where the $N$ is the test data size used for evaluation. A lower ECE indicates better calibration.

\paragraph{Temperature Scaling}
Temperature scaling\,\citep{guo2017calibration} is a widely used post-hoc calibration technique that assumes a validation set. The classifier's output logits are scaled by a temperature value $T>0$. 
\begin{gather}
\label{eq:TS_loss}
T^* = \argmin_{T} \mathcal{L}_{CE}(\mathbf{x}, y ; \theta, T)\\
\mathcal{L}_{CE}(\mathbf{x}, y ; \theta, T) = -\frac{1}{N} \sum_{i=1}^N  \sum_{k=1}^K y_{i, k} \frac{e^{z_{i,k}/T}}{\sum_{j=1}^K e^{z_{i,j}/T}} 
\end{gather}
If $T$ is large, the output probability distribution would be flattened into a uniform distribution; a small $T$ would sharpen the probability distribution instead. The optimal temperature is determined as the temperature that minimizes the cross-entropy loss on a validation set.

\paragraph{Adversarial Perturbation}
Adversarial perturbations add adversarial noise to input data to intentionally trigger misclassification. The Fast Gradient Sign Method (FGSM) \citep{goodfellow2014explaining} is an effective method for generating such perturbations. FGSM generates noise that maximizes the model's loss value for the given data and adds this noise to the data. FGSM perturbs a datapoint $\mathbf{x}$ in two ways, depending on whether a target class for misclassification is specified:
\begin{align}
\label{eq:FGSM_far}
\mathbf{x}^{adv} &= \mathbf{x} + \epsilon \text{ sign}(\nabla_{\mathbf{x}}\mathcal{L}_{CE}(\mathbf{x},y;\theta)) \\
\label{eq:FGSM_close}
\mathbf{x}^{adv} &= \mathbf{x} - \epsilon \text{ sign}(\nabla_{\mathbf{x}}\mathcal{L}_{CE}(\mathbf{x},y';\theta))
\end{align}
where $\epsilon$ is the magnitude of the perturbation, sign($\cdot$) is the sign function, $y$ is the true label, and $y'$ is the target class different from $y$ to induce a misclassification.

\section{Temperature Bias in CIL}
\label{sec:bias}
We investigate the calibration bias of temperature scaling when only new-task data is available for validation in class-incremental learning. While existing variants of temperature scaling address challenging scenarios like domain or distribution shifts with constrained validation sets, they still assume access to information from all classes via the validation set. In our setting, despite a consistent domain (image data), the validation set exclusively contains data from the new task, excluding any data from old tasks.

To verify the effectiveness of temperature scaling in this setting, we conduct experiments on two real-world image datasets: CIFAR-100 and Tiny-ImageNet. For CIFAR-100, we divide the dataset into 10 incremental tasks with 10 classes per task and train a 32-layer ResNet\,\citep{he2016deep} model. For Tiny-ImageNet, we train a ResNet-18 model with 10 tasks and 20 classes per task. In both cases, we employ Experience Replay (ER)\,\citep{chaudhry2019tiny} as our base learning technique.

\begin{figure}[t]
\centering
\begin{subfigure}{0.48\columnwidth}
\includegraphics[width=\columnwidth]{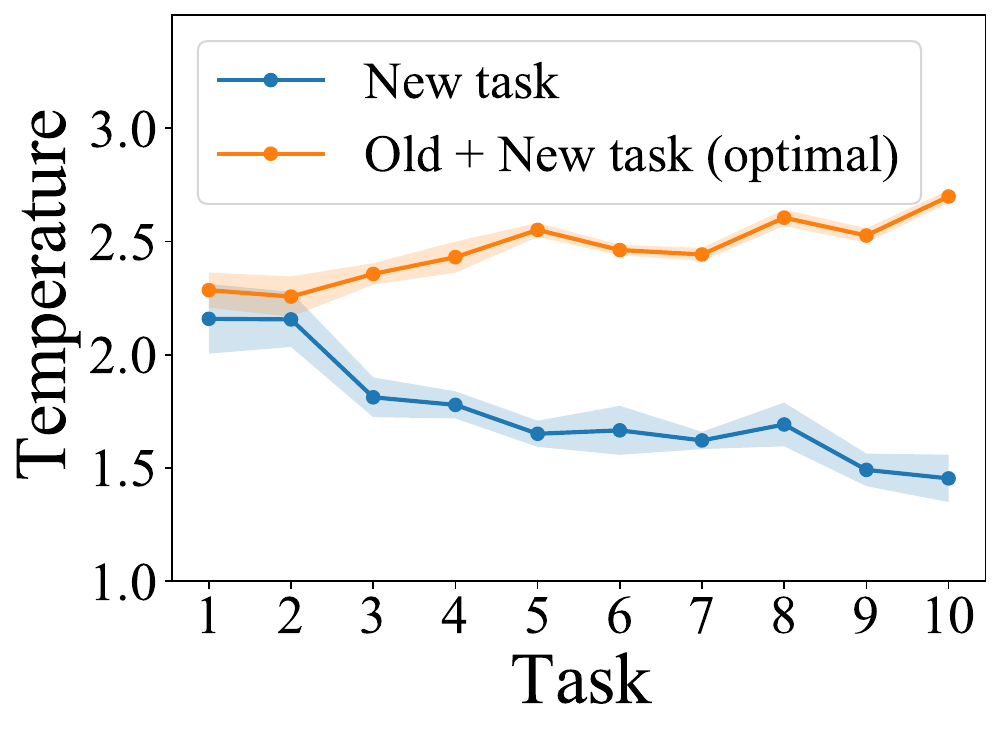}
\label{fig:temp_bias_cifar100}
\end{subfigure}
\begin{subfigure}{0.48\columnwidth}
\includegraphics[width=\columnwidth]{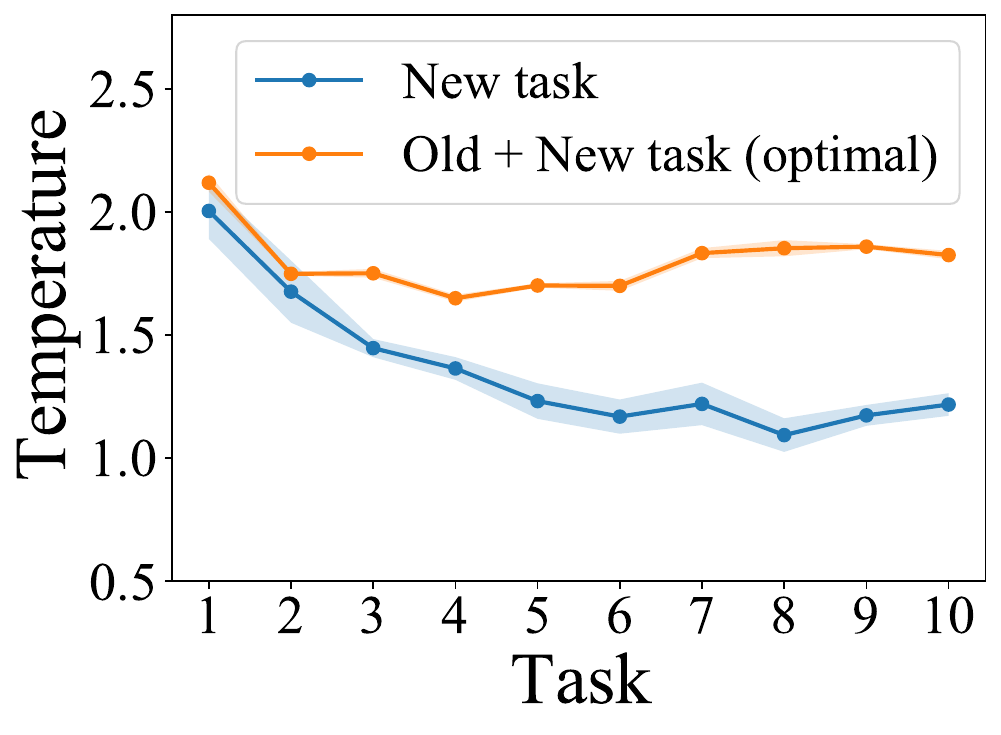}
\label{fig:temp_bias_tinyimagenet}
\end{subfigure}
\vskip -0.25in
\caption{Temperatures after training each task using ER method on the CIFAR-100 (\textit{left}), and Tiny-ImageNet (\textit{right}) datasets. For each task, we show two different temperatures: (1) optimized on a validation set containing only new-task data, and (2) the optimal temperature obtained on the test set across all tasks.}
\label{fig:temp_bias}
\vskip -0.1in
\end{figure}

After training, we optimize temperatures using two different sets: (1) a validation set containing data only from the new task, and (2) a test set containing data from all tasks (optimal). As shown in Figure~\ref{fig:temp_bias}, the gap between these optimized temperatures increases as the tasks progress. This observation suggests that using temperature optimized solely on new task validation data leads to poorer calibration performance on the actual test set, which includes data from both old and new tasks. This comparison highlights the potential bias when using only new-task data for temperature optimization.

\section{Method}
In this section, we introduce \method{}, a novel post-hoc calibration method specifically designed for class-incremental learning when a validation
set only from the new task is available. \method{} addresses the challenge of limited access to old-task data by leveraging adversarially perturbed exemplars. We propose an adversarial perturbation approach composed of perturbation direction and magnitude.
While one may use other augmentation techniques like Mixup for post-hoc calibration, adversarial perturbation is a more effective method to use because we need transformed old-task exemplars with degraded accuracy for calibration and adversarial perturbation serves the purpose directly. The full \method{} algorithm is also presented in Section~\ref{sec:overall_algorithm}.

\subsection{Perturbation Direction Policy}
\begin{figure}[t]
\centering
\begin{subfigure}{0.48\columnwidth}
\includegraphics[width=\columnwidth]{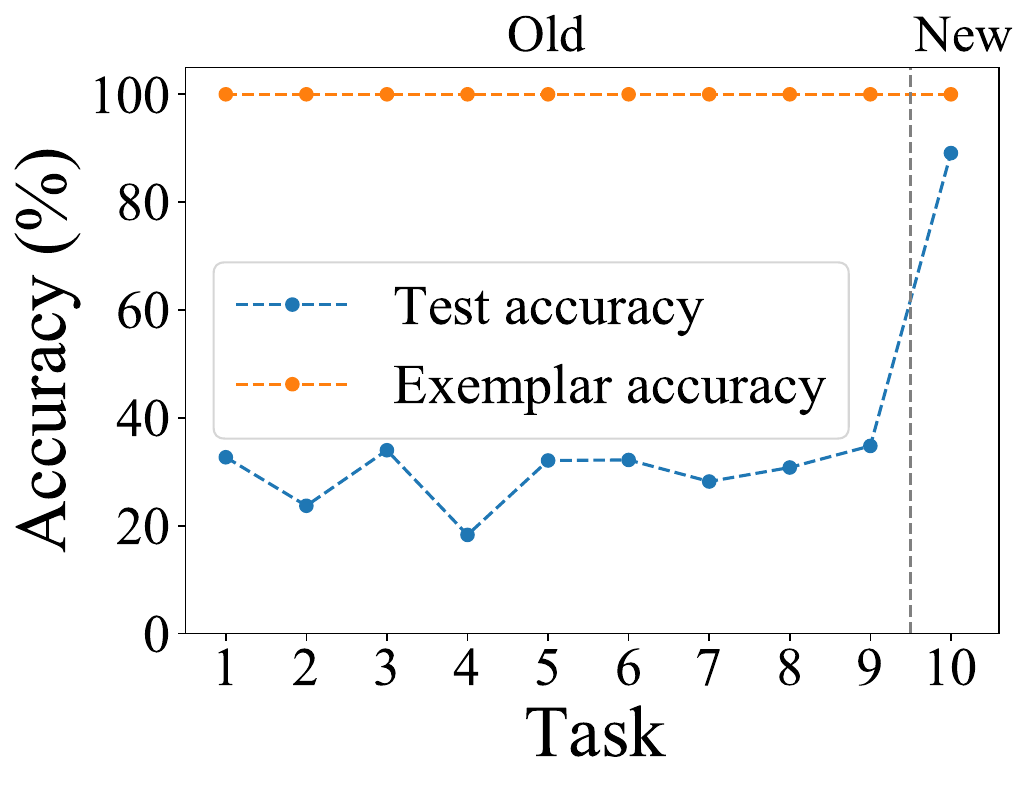}
% \caption{}
\label{fig:acc_hist_er}
\end{subfigure}
\begin{subfigure}{0.48\columnwidth}
\includegraphics[width=\columnwidth]{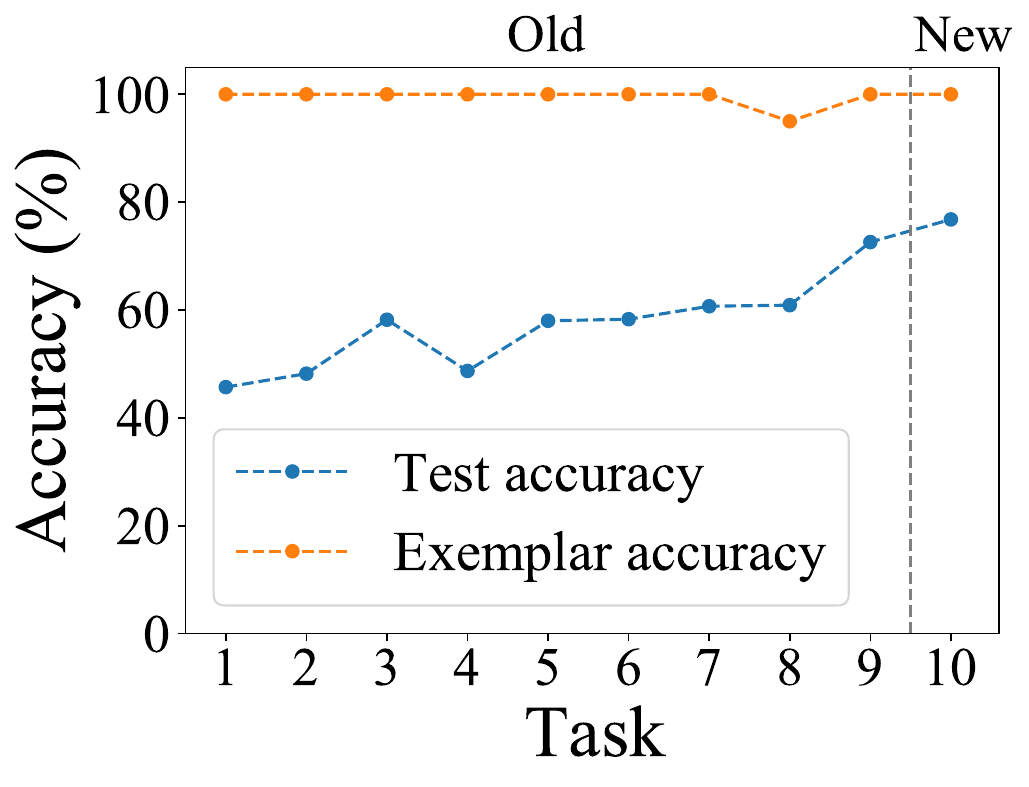}
% \caption{}
\label{fig:acc_hist_der}
\end{subfigure}
\vskip -0.2in
\caption{Task-wise accuracy comparison after completing training on the final (10th) task of CIFAR-100 dataset using Experience Replay (ER) (\textit{left}), and Dynamically Expandable Representation (DER)\,\citep{yan2021dynamically} (\textit{right}).}
\label{fig:acc_hist}
\vskip -0.2in
\end{figure}

To address the absence of a validation set from old tasks, we leverage exemplars from memory to gain information about old tasks. However, directly using these exemplars for temperature tuning leads to overfitting and inaccurate temperature estimates. This is because the model has already encountered these exemplars during training and tends to overfit on this limited data, resulting in high-confidence outputs and near-perfect accuracy. In fact, as shown in Figure~\ref{fig:acc_hist}, the model accuracy on exemplars is almost 100\% on the CIFAR-100, regardless of whether the learning technique is Experience Replay (ER) or Dynamically Expandable Representation (DER)\,\citep{yan2021dynamically}, which preserves the old task accuracy using a dynamic architecture.

This near-perfect accuracy disrupts the temperature optimization process, as described in Equation~\ref{eq:TS_loss}. Temperature optimization aims to find the temperature that minimizes the cross-entropy loss. When most predictions are correct, this process tends to lower the temperature, pushing confidence levels closer to 1. Consequently, using exemplars directly results in an abnormally low temperature, hindering calibration performance on the test set.

Thus, we perturb the exemplars adversarially triggering mispredictions to reduce overfitting for temperature optimization. Each perturbation of an exemplar has a direction and a magnitude. All perturbations have the same magnitude, which is configured with the parameter $\epsilon_{adv}$.
While we could use separate magnitude parameters for old and new tasks, determining the right magnitude for the old tasks is not easy without any validation data to tell us how the magnitude influences the temperature optimization. Thus we only employ a single magnitude value that is established using the validation data from the new tasks. We explain how to derive the magnitude in Section~\ref{sec:mag_search}.

With the single magnitude, we perturb exemplars more strongly for old tasks than for the new task by adjusting the perturbation direction via target class selection. For exemplars from old tasks, we select the easiest target classes to induce errors, while for new task exemplars, we select the hardest target classes. This idea is based on the tendency that the accuracy of old tasks is lower than that of the new task as shown in Figure~\ref{fig:acc_hist}. The results show that models after training the 10th task tend to perform better on new tasks, where more training data is available. 

\begin{figure}[t]
\centering
\includegraphics[width=0.85\columnwidth]{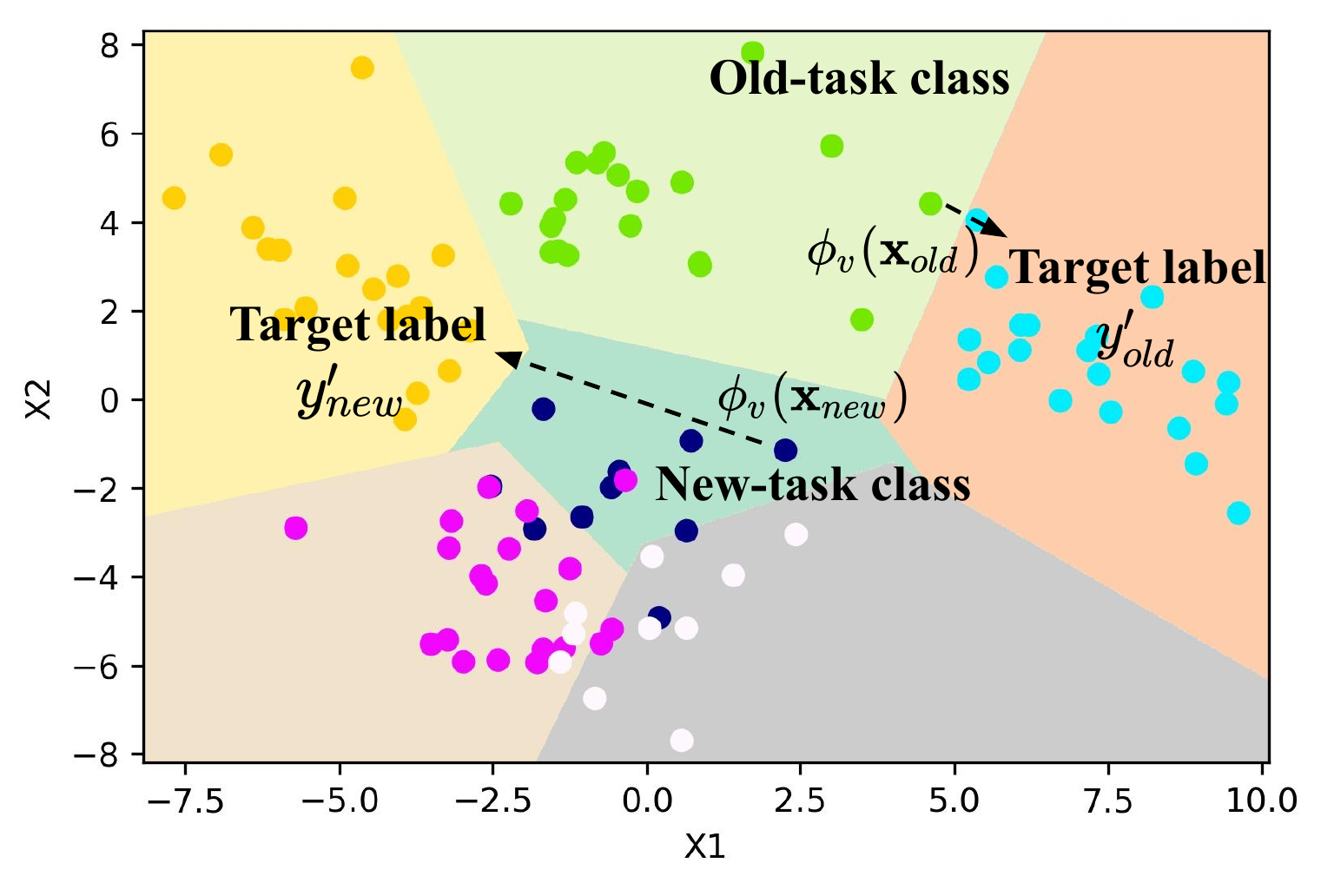}
\vskip -0.1in
\caption{A perturbation direction policy of \method{} visualized on a two-dimensional feature space with decision boundaries. For old-task data, the target class is selected based on the closest distance in feature space, while for the new-task data, the farthest class is selected based on maximum distance in the feature space.}
\label{fig:feature_space}
\vskip -0.15in
\end{figure}

We use feature distance within the feature space as an indicator of how easily an exemplar can cause mispredictions for different target classes, where the farthest class represents the most difficult target, and the nearest class represents the easiest target as represented in Figure~\ref{fig:feature_space}. To measure the feature distance, we decompose the classifier $f_{\theta}$ as $f_{\theta}(\cdot) = g_w(\phi_v(\cdot))$, where $\theta = \{w, v\}$, $\phi_v(\cdot)$ is the feature extractor with parameters $v$, and $g_w(\cdot)$ is the classification layer with parameters $w$. For each exemplar $\mathbf{x}_e \in \mathcal{M}_t$, where $\mathcal{M}_t$ contains exemplars from both old and new tasks, we compute the L2 distance between its feature $\phi_v(\mathbf{x}_e)$ and the mean of features for each class $\mu_c = \frac{1}{|X_c|} \sum_{\mathbf{x}_e \in X_c} \phi_v(\mathbf{x}_e)$, where $X_c$ is the set of exemplars of each class $c \in \{1, ..., C_{t, old}+C_t\}$. The target label $y'_e$ of data point $(\mathbf{x}_e, y_e)$ is then defined as follows:
\begin{equation}
y'_e = 
\begin{cases} 
\displaystyle\argmax_{c, c \neq y_e} \|\phi_v(\mathbf{x}_e) - \mu_c\| & \text{if } y_e \in \mathcal{Y}_t \\
\displaystyle\argmin_{c, c \neq y_e} \|\phi_v(\mathbf{x}_e) - \mu_c\| & \text{if } y_e \notin \mathcal{Y}_t
\end{cases}
\end{equation}
\vskip -0.15in
\begin{equation*}
\text{where } \mu_c = \frac{1}{|X_c|}\sum_{\mathbf{x}_e \in X_c} \phi_v(\mathbf{x}_e)
\end{equation*}

Then, we perturb exemplars to force the model directly to predict to the target class by using Equation~\ref{eq:FGSM_close}. This approach minimizes the loss concerning a chosen target class, thus actively pushing the model to misclassify the perturbed exemplars intentionally. We denote the resulting perturbed exemplar set as $\mathcal{M}^{adv}_t$. The optimized temperature, $T_{adv}$, is then determined by minimizing the cross-entropy loss on these adversarially perturbed exemplars as follows:
\begin{align} 
\label{eq:perturb}
T_{adv} = \argmin_{T} \mathcal{L}_{CE}(\mathbf{x}^{adv}_e, y_e; \theta, T) \\
\mathbf{x}^{adv}_e = \mathbf{x}_e - \epsilon \text{ sign}(\nabla_{\mathbf{x}_e}\mathcal{L}_{CE}(\mathbf{x}_e, y'_e ; \theta))
\end{align}
where $\mathbf{x}^{adv}_e$ represents the perturbed exemplar data, $y'_e$ is the target class of perturbation, $y_e$ is the original label of $\mathbf{x}_e$, and $\epsilon$ is the magnitude of perturbation.

We provide supporting experiments on why it makes sense for \method{} to perturb old-task exemplars towards closer classes to match the lower real accuracy. Using \method{} with ER on CIFAR-100 and comparing results after training up to the 5th and 10th tasks (see Section~\ref{sec:exp} for the setup details), the accuracy gap between old and new tasks increases from 38\% to 60\%. At the same time, \method{}'s misprediction rate gap increases from 17.8\% to 43.7\%, which is proportional to the accuracy gap increase.

\subsection{Perturbation Magnitude Search}
\label{sec:mag_search}
We determine the perturbation magnitude parameter $\epsilon_{adv}$ using a new-task validation set, since existing works\,\citep{goodfellow2014explaining, carlini2017towards} for adversarial perturbation determine the magnitude heuristically as a hyperparameter. $\epsilon_{adv}$ is selected to yield a temperature on perturbed exemplars that matches the optimal temperature from the new-task validation set.

Initially, we determine a target temperature by minimizing the cross-entropy loss on $\mathcal{D}_{t, valid}$, a validation set from the new task. We find the magnitude of perturbation, $\epsilon_{adv}$ that results in this target temperature when applied to perturbations on $\mathcal{M}_{t, new}$, the exemplar set from the new task. Note that $\mathcal{M}_{t, new}$ is sampled from $\mathcal{D}_{t, train}$ after training, which itself is obtained by splitting the whole new-task data $\mathcal{D}_t$ into a training set, $\mathcal{D}_{t, train}$, and a validation set, $\mathcal{D}_{t, valid}$.

Since the increase in $\epsilon$ leads to more incorrect predictions by the model, resulting in higher optimized temperatures, we adopt a binary search algorithm to identify the $\epsilon_{adv}$. We denote the perturbed new task exemplar set with magnitude $\epsilon$ as $\mathcal{M}^{\epsilon}_{t, new}$. The $\epsilon_{adv}$ value is determined as follows:
\begin{align}
\label{eq:mag_search}
\epsilon_{adv} = \argmin_{\epsilon} \left\| T_{{target}} - T_{adv}(\epsilon) \right\| \\
T_{{target}} = \argmin_T \mathcal{L}_{{CE}}(\mathbf{x}_v, y_v; \theta, T)  \\
T_{adv}(\epsilon) = \argmin_T \mathcal{L}_{{CE}}(\mathbf{x}^{adv}_e, y_e; \theta, T)
\end{align}
where $(\mathbf{x}_v, y_v) \in \mathcal{D}_{t, {valid}}$ and  $(\mathbf{x}^{adv}_e, y_e) \in \mathcal{M}^{\epsilon}_{t, {new}}$.

\begin{algorithm}[t]
\SetKwInput{Input}{Input}
\SetKwInOut{Output}{Output}
\SetNoFillComment
\Input{Exemplar set $\mathcal{M}_{t}$, validation set from new task $\mathcal{D}_{t, valid}$, model parameters $\theta = \{w, v\}$, tolerance $\delta$}
/* At the $t$-th incremental task */ \\
$T_{target} = \texttt{TempOpt}(\mathcal{D}_{t, valid}, \theta)$ \\
$\mu = \{\mu_1, \mu_2, ..., \mu_{C_{t, old} + C_t}\}$, where $\mu_c = \frac{1}{|X_c|} \sum_{\mathbf{x}_e \in X_c} \phi_v(\mathbf{x}_e)$\\
$\mathcal{M}_{t, new} = \{(\mathbf{x}_e, y_e) \mid (\mathbf{x}_e, y_e) \in \mathcal{M}_t, y_e \in \mathcal{Y}_t\}$ \\
$\epsilon_{adv} = \texttt{MagSearch}(\theta, \mathcal{M}_{t, new}, T_{target}, \mu, \delta)$ \\
$\mathcal{M}^{\epsilon_{adv}}_{t} = \{\ \}$ \\
\For {$(\mathbf{x}_e, y_e) \in \mathcal{M}_{t}$} {
    $y'_e = 
    \begin{cases} 
    \displaystyle\argmax_{c : \mu_c \in \mu, c \neq y_e} \|\phi_v(\mathbf{x}_e) - \mu_c\| & \text{if } y_e \in \mathcal{Y}_t \\
    \displaystyle\argmin_{c : \mu_c \in \mu, c \neq y_e} \|\phi_v(\mathbf{x}_e) - \mu_c\| & \text{if } y_e \notin \mathcal{Y}_t
    \end{cases}$ \\
    $\mathbf{x}^{adv}_e = \mathbf{x}_e - \epsilon_{adv} \text{ sign}(\nabla_{\mathbf{x}_e}\mathcal{L}_{CE}(\mathbf{x}_e, y'_e ; \theta))$ \\
    $\mathcal{M}^{\epsilon_{adv}}_{t} \leftarrow \mathcal{M}^{\epsilon_{adv}}_{t} \cup \{(\mathbf{x}^{adv}_e, y_e)\}$
}
$T_{adv} = \texttt{TempOpt}(\mathcal{M}^{\epsilon_{adv}}_{t}, \theta)$ \\
\Output{$T_{adv}$}
\caption{The \method{} algorithm.}
\label{alg:T-CIL}
% \vskip -0.2in
\end{algorithm}

\subsection{Overall Algorithm}
\label{sec:overall_algorithm}
Algorithm~\ref{alg:T-CIL} presents the overall procedure of \method{}, which is designed for class-incremental learning scenarios with multiple tasks. Initially, we determine the target temperature on $\mathcal{D}_{t, valid}$ using Algorithm~\ref{alg:temp_opt} (\texttt{TempOpt}, detailed in Appendix) (Line 2). We compute the mean of features for all classes, including both old and new task classes (Line 3). We sample the new task exemplar set, $\mathcal{M}_{t, new}$ from $\mathcal{M}_t$ (Line 4). Subsequently, we employ the binary search algorithm described in Algorithm~\ref{alg:binary_search} (\texttt{MagSearch}, detailed in Appendix) to determine $\epsilon_{adv}$ on $\mathcal{M}_{t, new}$ (Line 5). Using $\epsilon_{adv}$ and our perturbation direction policy, we perturb all exemplars (Lines 6--10). Finally, we optimize the temperature on the perturbed exemplar set using Algorithm~\ref{alg:temp_opt} (Line 11). This optimized temperature will be used for scaling the output logits of the model. 

We also analyze the computational complexity of \method{}. Typical class-incremental learning methods have a complexity of $O(T(N_{\text{new}}+M))$, where $T$ is the number of tasks, $N_{\text{new}}$ is the number of new task data points, and $M$ is the memory size. \method{} itself has a complexity of $O(M)$ as its temperature optimization, feature means calculation, perturbation magnitude search, and memory update takes $O(M)$ time (see Appendix for details). Therefore, when combining \method{} with class-incremental learning methods, the overall  complexity remains asymptotically unchanged as $M$ is fixed and significantly smaller than $N_{\text{new}}$.

\begin{table*}[h]
\centering
\caption{\method{} performance compared to the five baseline methods on three datasets.}
\small
\vskip -0.05in
\begin{tabular}{l|ccc|ccc|ccc}
\toprule
{} & \multicolumn{3}{c|}{CIFAR-10} & \multicolumn{3}{c|}{CIFAR-100} & \multicolumn{3}{c}{Tiny-ImageNet} \\
\cmidrule{1-10}
{Method} & {Acc. ($\uparrow$)} & {ECE ($\downarrow$)} & {AECE ($\downarrow$)} & {Acc. ($\uparrow$)} & {ECE ($\downarrow$)} & {AECE ($\downarrow$)} & {Acc. ($\uparrow$)} & {ECE ($\downarrow$)} & {AECE ($\downarrow$)}  \\
\midrule
{Vanilla} & {65.61}\tiny{$\pm$ 0.49} & {28.16}\tiny{$\pm$ 0.33} & {28.12}\tiny{$\pm$ 0.32} & {56.51}\tiny{$\pm$ 0.37} & {27.66}\tiny{$\pm$ 0.29} & {27.64}\tiny{$\pm$ 0.28} & {31.95}\tiny{$\pm$ 0.54} & {32.56}\tiny{$\pm$ 0.36} & {32.55}\tiny{$\pm$ 0.35} \\
\midrule
{TS} & {65.86}\tiny{$\pm$ 0.09} & {23.97}\tiny{$\pm$ 0.82} & {23.92}\tiny{$\pm$ 0.81} & {56.25}\tiny{$\pm$ 0.62} & {16.28}\tiny{$\pm$ 0.32} & {16.22}\tiny{$\pm$ 0.37} & {31.48}\tiny{$\pm$ 0.39} & {19.44}\tiny{$\pm$ 0.75} & {19.44}\tiny{$\pm$ 0.75} \\
{ETS} & {65.86}\tiny{$\pm$ 0.09} & {22.46}\tiny{$\pm$ 1.20} & {22.39}\tiny{$\pm$ 1.21} & {56.25}\tiny{$\pm$ 0.62} & {16.83}\tiny{$\pm$ 0.47} & {16.80}\tiny{$\pm$ 0.49} & {31.48}\tiny{$\pm$ 0.39} & {19.81}\tiny{$\pm$ 0.65} & {19.83}\tiny{$\pm$ 0.65} \\
{IRM} & {65.86}\tiny{$\pm$ 0.09} & {22.09}\tiny{$\pm$ 1.73} & {21.83}\tiny{$\pm$ 1.73} & {56.25}\tiny{$\pm$ 0.62} & {17.50}\tiny{$\pm$ 0.64} & {17.39}\tiny{$\pm$ 0.61} & {31.48}\tiny{$\pm$ 0.39} & {19.77}\tiny{$\pm$ 0.72} & {19.73}\tiny{$\pm$ 0.78} \\
{PerturbTS} & {n/a} & {n/a} & {n/a} & {56.25}\tiny{$\pm$ 0.62} & {16.49}\tiny{$\pm$ 2.19} & {16.49}\tiny{$\pm$ 2.18} & {31.48}\tiny{$\pm$ 0.39} & {10.60}\tiny{$\pm$ 0.60} & {10.58}\tiny{$\pm$ 0.61} \\
{\textbf{\method{}}} & {65.86}\tiny{$\pm$ 0.09} & \textbf{{17.70}\tiny{$\pm$ 2.60}} & \textbf{{17.64}\tiny{$\pm$ 2.59}} & {56.25}\tiny{$\pm$ 0.62} & \textbf{{5.74}\tiny{$\pm$ 0.53}} & \textbf{{5.75}\tiny{$\pm$ 0.50}} & {31.48}\tiny{$\pm$ 0.39} & \textbf{{8.12}\tiny{$\pm$ 0.38}} & \textbf{{8.12}\tiny{$\pm$ 0.41}} \\
\bottomrule
\end{tabular}
\label{tbl:baselines}
\end{table*}

\begin{table*}[h]
\centering
\caption{\method{} performance combined with four existing class-incremental learning techniques on three datasets.}
\vskip -0.05in
\small
\begin{tabular}{l|ccc|ccc|ccc}
\toprule
{} & \multicolumn{3}{c|}{CIFAR-10} & \multicolumn{3}{c|}{CIFAR-100} & \multicolumn{3}{c}{Tiny-ImageNet} \\
\cmidrule{1-10}
{Method} & {Acc. ($\uparrow$)} & {ECE ($\downarrow$)} & {AECE ($\downarrow$)} & {Acc. ($\uparrow$)} & {ECE ($\downarrow$)} & {AECE ($\downarrow$)} & {Acc. ($\uparrow$)} & {ECE ($\downarrow$)} & {AECE ($\downarrow$)} \\
\midrule
{ER} & {65.61}\tiny{$\pm$ 0.49} & {28.16}\tiny{$\pm$ 0.33} & {28.12}\tiny{$\pm$ 0.32} & {56.51}\tiny{$\pm$ 0.37} & {27.66}\tiny{$\pm$ 0.29} & {27.64}\tiny{$\pm$ 0.28} & {31.95}\tiny{$\pm$ 0.54} & {32.56}\tiny{$\pm$ 0.36} & {32.55}\tiny{$\pm$ 0.35} \\
{ER+\method{}} & {65.86}\tiny{$\pm$ 0.09} & \textbf{{17.70}\tiny{$\pm$ 2.60}} & \textbf{{17.64}\tiny{$\pm$ 2.59}} & {56.25}\tiny{$\pm$ 0.62} & \textbf{{5.74}\tiny{$\pm $0.53}} & \textbf{{5.75}\tiny{$\pm$ 0.50}} & {31.48}\tiny{$\pm$ 0.39} & \textbf{{8.12}\tiny{$\pm$ 0.38}} & \textbf{{8.12}\tiny{$\pm$ 0.41}} \\
\cmidrule{1-10}
{EEIL} & {77.67}\tiny{$\pm$ 0.74} & {15.48}\tiny{$\pm$ 0.75} & {15.45}\tiny{$\pm$ 0.74} & {60.61}\tiny{$\pm$ 0.33} & {21.96}\tiny{$\pm$ 0.29} & {21.94}\tiny{$\pm$ 0.28} & {37.44}\tiny{$\pm$ 0.85} & {29.69}\tiny{$\pm$ 0.36} & {29.68}\tiny{$\pm$ 0.36} \\
{EEIL+\method{}} & {76.91}\tiny{$\pm$ 1.27} & \textbf{{10.49}\tiny{$\pm$ 2.34}} & \textbf{{10.46}\tiny{$\pm$ 2.32}} & {60.74}\tiny{$\pm$ 0.37} & \textbf{{10.30}\tiny{$\pm$1.10}} & \textbf{{10.22}\tiny{$\pm$ 1.06}} & {37.16}\tiny{$\pm$ 0.91} & \textbf{{15.58}\tiny{$\pm$ 0.76}} & \textbf{{15.56}\tiny{$\pm$ 0.76}} \\
\cmidrule{1-10}
{WA} & {73.06}\tiny{$\pm$ 0.57} & {19.10}\tiny{$\pm$ 0.50} & {19.07}\tiny{$\pm$ 0.51} & {64.34}\tiny{$\pm$ 0.40} & {8.89}\tiny{$\pm$ 0.64} & {8.86}\tiny{$\pm$ 0.63} & {39.66}\tiny{$\pm$ 0.88} & \textbf{{10.97}\tiny{$\pm$ 0.41}} & \textbf{{10.96}\tiny{$\pm$ 0.43}} \\
{WA+\method{}} & {72.75}\tiny{$\pm$ 0.47} & \textbf{{15.61}\tiny{$\pm$ 0.23}} & \textbf{{15.58}\tiny{$\pm$ 0.22}} & {64.02}\tiny{$\pm$ 0.06} & \textbf{{3.87}\tiny{$\pm$ 0.52}} & \textbf{{3.84}\tiny{$\pm$ 0.55}} & {38.59}\tiny{$\pm$ 0.44} & {11.43}\tiny{$\pm$ 1.00} & {11.46}\tiny{$\pm$ 1.04} \\
\cmidrule{1-10}
{DER} & {74.53}\tiny{$\pm$  0.48} & {21.81}\tiny{$\pm$ 0.46} & {21.78}\tiny{$\pm$ 0.47} & {69.98}\tiny{$\pm$ 0.69} & {22.38}\tiny{$\pm$ 0.37} & {22.35}\tiny{$\pm$ 0.35} & {46.62}\tiny{$\pm$ 2.84} & {39.00}\tiny{$\pm$ 1.72} & {38.99}\tiny{$\pm$ 1.72} \\
{DER+\method{}} & {74.93}\tiny{$\pm$ 0.35} & \textbf{{12.70}\tiny{$\pm$ 1.35}} & \textbf{{12.68}\tiny{$\pm$ 1.34}} & {69.98}\tiny{$\pm$ 0.58} & \textbf{{4.37}\tiny{$\pm$ 0.59}} & \textbf{{4.34}\tiny{$\pm$ 0.60}} & {47.79}\tiny{$\pm$ 0.47} & \textbf{{6.91}\tiny{$\pm$ 0.86}} & \textbf{{6.90}\tiny{$\pm$ 0.84}} \\
\bottomrule
\end{tabular}
\label{tbl:compatibility}
\vskip -0.1in
\end{table*}

\section{Experiments}
\label{sec:exp}
We provide experimental results for \method{}, evaluating the calibration performance of classifiers in a class-incremental learning setting. We report the results with the mean and the standard deviation ($\pm$ in tables) for five random seeds. We use PyTorch\,\citep{paszke2019pytorch} with NVIDIA GeForce RTX 4090 GPUs for all experiments.

\paragraph{Metrics}
We report the top-1 Accuracy, Expected Calibration Error (ECE), and Adaptive Expected Calibration Error (AECE) in percentages. ECE\,\citep{naeini2015obtaining} is defined as $\sum^B_{i=1} \frac{|B_i|}{N} |acc(B_i) - conf(B_i)|$, where $acc(B_i)$ is the average accuracy, $conf(B_i)$ is the average confidence of each bin $B_i$, and $N$ is the number of test data. We use $B=10$ bins. AECE\,\citep{mukhoti2020calibrating} is an another metric for calibration performance evaluation with uniform number of samples for each bin. AECE is also defined as $\sum^B_{i=1} \frac{|B_i|}{N} |acc(B_i) - conf(B_i)|$, but each bin contains $\frac{N}{B}$ samples uniformly, sorted by ascending order of the confidence level. Similar to how taking the average accuracy across tasks is a main metric for class-incremental learning, we take the average calibration error such as average ECE across tasks as a calibration performance metric for class-incremental learning.

\paragraph{Datasets}
We evaluate our method on three widely used benchmarks for class-incremental learning: CIFAR-10, CIFAR-100, and Tiny-ImageNet. CIFAR-10 and CIFAR-100~\citep{krizhevsky2009learning} share similar characteristics, both consisting of 32 × 32 RGB images. CIFAR-10 consists of 10 object classes with 5,000 training and 1,000 test images per class, totaling 50,000 training and 10,000 test images. Similarly, CIFAR-100 contains 100 classes, but with fewer images per class: 500 for training and 100 for testing, maintaining the same total of 50,000 training and 10,000 test images. Tiny-ImageNet~\citep{le2015tiny}, a compact version of ImageNet ILSVRC 2021\,\citep{russakovsky2015imagenet}, contains 200 classes of 64 × 64 RGB images, with 500 training, and 50 test images per class.

\paragraph{Experimental Settings}
For CIFAR-10, we employ a 32-layer ResNet architecture (ResNet-32) \citep{he2016deep}. We divide the dataset into 5 incremental tasks, with 2 classes per task. We maintain a memory of 200 samples and use a batch size of 128. For CIFAR-100, we also utilize ResNet-32 and construct the learning with 10 incremental tasks, each containing 10 classes. The memory size is 2,000 samples and the batch size is 128. For Tiny-ImageNet, we adopt ResNet-18 and organize the learning into 10 incremental tasks with 20 classes per task. The memory size is 4,000 samples, and we use the batch size of 256. 

Across all datasets, we apply data augmentation techniques including random cropping and horizontal flipping. We utilize maximum memory storage from the initial task to the last task. For all experiments, we train a model for 200 epochs using the Adam\,\citep{kingma2014adam} optimizer with a weight decay of 0.0002 and use an initial learning rate of 0.1, which is divided by a factor of 10 at epochs 100 and 150. 

\paragraph{Baselines}
We compare \method{} to existing post-hoc calibration methods adaptable to class-incremental learning settings as baselines: 1) \textit{Vanilla} is a class-incremental learning technique without any calibration method; 2) \textit{TS}\,\citep{guo2017calibration} is a temperature scaling; 3) \textit{ETS}\,\citep{zhang2020mix} is an ensemble temperature scaling balancing between calibrated and uncalibrated logits; 4) \textit{IRM}\,\citep{zhang2020mix} is a multi-class isotonic regression method; 5) \textit{PerturbTS}\,\citep{tomani2021post} is an improved temperature scaling method using a perturbed validation set. Since we do not have a validation set from old tasks, \method{} and the baselines above only use a validation set from the new task. 

\vskip -0.1in

\subsection{Calibration Results}
We compare the overall calibration performance of \method{} with the five baselines on three datasets as shown in Table~\ref{tbl:baselines}. We use ER\,\citep{chaudhry2019tiny} as a base class-incremental learning method. For post-hoc calibration baselines except \textit{Vanilla}, we use 500 samples as a validation set for the new task.
While simple calibration methods can improve model calibration in class-incremental learning, \method{} achieves substantially better calibration performance across three datasets. Specifically, on the CIFAR-100, \method{} shows remarkable improvement with a 65.2\% reduction in calibration error compared to the best post-hoc calibration baselines. The improvements are also significant for other datasets, with 20.0\% and 23.4\% reductions in calibration error on the CIFAR-10 and Tiny-ImageNet, respectively, compared to the best performing baselines. The `n/a' in Table~\ref{tbl:baselines} for \textit{PerturbTS} on the CIFAR-10 indicates that the method is not applicable to this dataset due to the limited number of classes per task (see details in the Appendix). 

\begin{table}[t]
\centering
\caption{\method{} performance compared to the three alternative perturbation direction policies on the CIFAR-100.}
\vskip -0.05in
\small
\begin{tabular}{l|ccc}
\toprule
{Method} & {Acc. ($\uparrow$)} & {ECE ($\downarrow$)} & {AECE ($\downarrow$)} \\
\midrule
{Random class} & {56.25}\tiny{$\pm$ 0.62} & {15.09}\tiny{$\pm$ 0.54} & {15.05}\tiny{$\pm$ 0.59} \\
{Closest class only} & {56.25}\tiny{$\pm$ 0.62} & {16.39}\tiny{$\pm$ 0.73} & {16.32}\tiny{$\pm$ 0.71} \\
{Farthest class only} & {56.25}\tiny{$\pm$ 0.62} & {14.52}\tiny{$\pm$ 0.45} & {14.49}\tiny{$\pm$ 0.46} \\
{\textbf{\method{}}} & {56.25}\tiny{$\pm$ 0.62} & \textbf{{5.74}\tiny{$\pm 0.53$}} & \textbf{{5.75}\tiny{$\pm$ 0.50}} \\
\bottomrule
\end{tabular}
\label{tbl:label_policy}
\vskip -0.1in
\end{table}

\subsection{Compatibility with CIL Techniques}
\label{sec:compatibility}
We evaluate the overall performance of \method{} when combining class-incremental learning techniques in Table~\ref{tbl:compatibility}.
Since \method{} is a post-hoc calibration method, \method{} can be integrated with any existing class-incremental learning technique alleviating catastrophic forgetting including ER\,\citep{chaudhry2019tiny}, EEIL\,\citep{castro2018end}, WA\,\citep{zhao2020maintaining}, and DER\,\citep{yan2021dynamically}. ER is a basic experience replay approach, EEIL and WA are based on knowledge distillation, and DER dynamically expands its architecture as the tasks progress. All methods utilize the memory. For details about experimental settings for each learning techniques including the new-task validation set size, please refer to the Appendix.

Consequently, \method{} improves the calibration performance across class-incremental learning techniques, which show poor calibration due to their focus on accuracy. While DER achieves the best accuracy on the CIFAR-100 and Tiny-ImageNet, it exhibits the highest calibration error among the techniques. When integrated with DER, \method{} reduces calibration error by 80.5\% and 82.3\% on these datasets, respectively. These substantial improvements across different techniques demonstrate \method{}'s versatility as a general calibration solution for class-incremental learning.
When combined with WA, \method{} shows varying performances across datasets. While achieving significant calibration improvements on the CIFAR-10 and CIFAR-100, it shows slightly higher calibration error than vanilla WA on the Tiny-ImageNet, likely due to WA's implicit logit scaling for the new task. Nevertheless, the overall benefits of \method{} across different techniques and datasets demonstrate its effectiveness as a general calibration solution. 

We also present the progression of ECE across tasks on CIFAR-100 and Tiny-ImageNet of vanilla, \method{}, and Optimal TS when combined with four class-incremental learning techniques in the Appendix. When combined with various class-incremental learning techniques, \method{} consistently demonstrates strong calibration performance across all tasks, with calibration errors approaching Optimal TS's ideal performance (see details in the Appendix).

\begin{figure}[t]
\centering
\includegraphics[width=0.85\columnwidth]{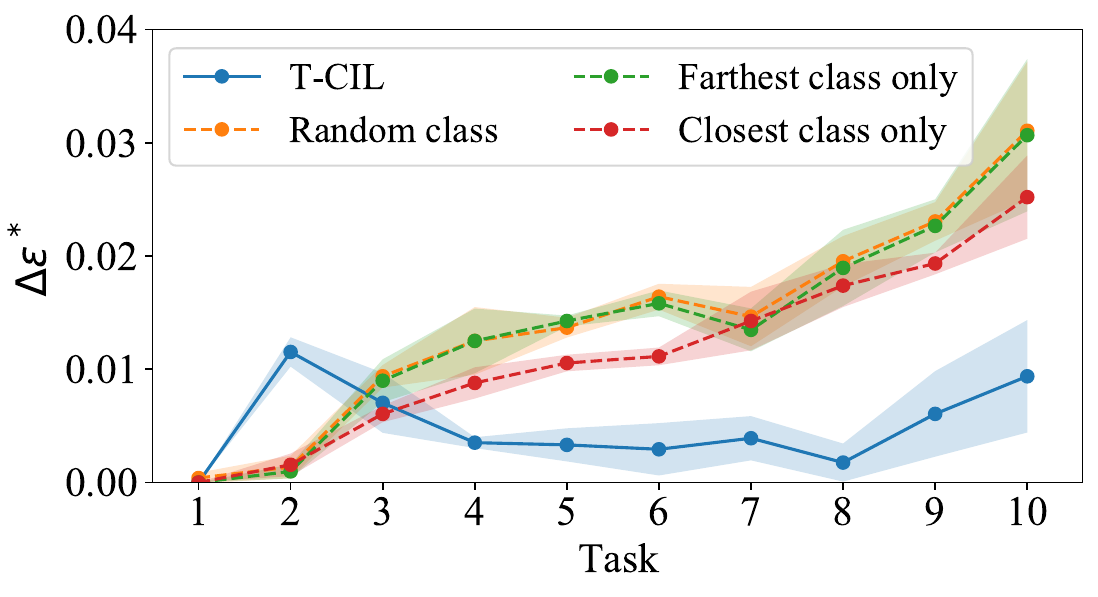}
\vskip -0.16in
\caption{The optimal perturbation magnitude differences ($\Delta\epsilon^*$) on the CIFAR-100 test sets after training each task, comparing the gap between old and new tasks versus new tasks only. Results are presented for the perturbation direction policies of \method{} and three alternative policies.}
\label{fig:epsilon_diff_trend}
\vskip -0.18in
\end{figure}

\subsection{Varying Perturbation Direction Policy}
To demonstrate the effectiveness of the perturbation direction policy of \method{}, we compare it against three possible policy types: (1) \textit{Random class}; (2) \textit{Closest class only}; and (3) \textit{Farthest class only}. ER is used as the base learning method. As shown in Table~\ref{tbl:label_policy}, \method{} is the only effective approach outperforming other policies that exhibit similar calibration performances with post-hoc calibration baselines. These results demonstrate that the perturbation direction policy of \method{} is essential for achieving effective calibration in class-incremental learning settings.

We demonstrate that our policy makes $\epsilon_{adv}$ determined from only the new-task validation set sufficient to calibrate both old and new tasks. We analyze the optimal perturbation magnitude differences (i.e., $\Delta\epsilon^* = |\epsilon^*_{old,new} - \epsilon^*_{new}|$) on the CIFAR-100, where $\epsilon^*_{old,new}$ and $\epsilon^*_{new}$ are searched from both old and new task test sets and only the new task test set, respectively. The optimal magnitude is determined when the temperature calculated on the perturbed exemplars matches the optimal temperature on that set. As shown in Figure~\ref{fig:epsilon_diff_trend}, our proposed perturbation direction policy maintains small magnitude differences consistently, while other policies show strictly increasing differences, as the tasks progress. This demonstrates that our method can effectively calibrate the model on both old and new tasks using a single $\epsilon_{adv}$ from only the new-task validation set.

\section{Conclusion}
We proposed \method{}, a novel and effective post-hoc calibration approach tailored to class-incremental learning settings. The existing post-hoc calibration methods are not designed to consider class-incremental learning settings, where validation data of old tasks is limited. \method{} leverages exemplars for calibration by applying adversarial perturbation to an exemplar set and optimizing the temperature on the perturbed set. The perturbation strategy consists of two key components: direction and magnitude. We determine the magnitude from the new-task validation set and strategically adjust the perturbation direction by guiding the model to misclassify old task samples into their closest classes while directing new task samples into their farthest classes, creating a balanced difficulty in misprediction. We conducted extensive experiments, which showed that \method{} significantly outperforms post-hoc calibration baselines. We also showed how \method{} is compatible with various existing learning techniques and improves their calibration performances.

{
    \small
    \bibliographystyle{ieeenat_fullname}
    \bibliography{main}

\begin{thebibliography}{47}
\providecommand{\natexlab}[1]{#1}
\providecommand{\url}[1]{\texttt{#1}}
\expandafter\ifx\csname urlstyle\endcsname\relax
  \providecommand{\doi}[1]{doi: #1}\else
  \providecommand{\doi}{doi: \begingroup \urlstyle{rm}\Url}\fi

\bibitem[Aljundi et~al.(2019)Aljundi, Belilovsky, Tuytelaars, Charlin, Caccia, Lin, and Page-Caccia]{aljundi2019online}
Rahaf Aljundi, Eugene Belilovsky, Tinne Tuytelaars, Laurent Charlin, Massimo Caccia, Min Lin, and Lucas Page-Caccia.
\newblock Online continual learning with maximal interfered retrieval.
\newblock \emph{Advances in Neural Information Processing Systems}, 32, 2019.

\bibitem[Carlini and Wagner(2017)]{carlini2017towards}
Nicholas Carlini and David Wagner.
\newblock Towards evaluating the robustness of neural networks.
\newblock In \emph{2017 ieee symposium on security and privacy (sp)}, pages 39--57. IEEE, 2017.

\bibitem[Castro et~al.(2018)Castro, Mar{\'\i}n-Jim{\'e}nez, Guil, Schmid, and Alahari]{castro2018end}
Francisco~M Castro, Manuel~J Mar{\'\i}n-Jim{\'e}nez, Nicol{\'a}s Guil, Cordelia Schmid, and Karteek Alahari.
\newblock End-to-end incremental learning.
\newblock In \emph{Proceedings of the European Conference on Computer Vision}, pages 233--248, 2018.

\bibitem[Cauwenberghs and Poggio(2000)]{cauwenberghs2000incremental}
Gert Cauwenberghs and Tomaso Poggio.
\newblock Incremental and decremental support vector machine learning.
\newblock \emph{Advances in Neural Information Processing Systems}, 13, 2000.

\bibitem[Chaudhry et~al.(2018)Chaudhry, Ranzato, Rohrbach, and Elhoseiny]{chaudhry2018efficient}
Arslan Chaudhry, Marc'Aurelio Ranzato, Marcus Rohrbach, and Mohamed Elhoseiny.
\newblock Efficient lifelong learning with a-gem.
\newblock \emph{arXiv preprint arXiv:1812.00420}, 2018.

\bibitem[Chaudhry et~al.(2019)Chaudhry, Rohrbach, Elhoseiny, Ajanthan, Dokania, Torr, and Ranzato]{chaudhry2019tiny}
Arslan Chaudhry, Marcus Rohrbach, Mohamed Elhoseiny, Thalaiyasingam Ajanthan, Puneet~K Dokania, Philip~HS Torr, and Marc'Aurelio Ranzato.
\newblock On tiny episodic memories in continual learning.
\newblock \emph{arXiv preprint arXiv:1902.10486}, 2019.

\bibitem[Chidambaram and Ge(2024)]{chidambaram2024limitations}
Muthu Chidambaram and Rong Ge.
\newblock On the limitations of temperature scaling for distributions with overlaps.
\newblock \emph{International Conference on Learning Representations}, 2024.

\bibitem[Choi et~al.(2024)Choi, Park, Han, Park, and Moon]{choi2024consistency}
Wonjeong Choi, Jungwuk Park, Dong-Jun Han, Younghyun Park, and Jaekyun Moon.
\newblock Consistency-guided temperature scaling using style and content information for out-of-domain calibration.
\newblock \emph{arXiv preprint arXiv:2402.15019}, 2024.

\bibitem[Ding et~al.(2021)Ding, Han, Liu, and Niethammer]{ding2021local}
Zhipeng Ding, Xu Han, Peirong Liu, and Marc Niethammer.
\newblock Local temperature scaling for probability calibration.
\newblock In \emph{Proceedings of the IEEE/CVF International Conference on Computer Vision}, pages 6889--6899, 2021.

\bibitem[Ghosh et~al.(2022)Ghosh, Schaaf, and Gormley]{ghosh2022adafocal}
Arindam Ghosh, Thomas Schaaf, and Matthew Gormley.
\newblock Adafocal: Calibration-aware adaptive focal loss.
\newblock \emph{Advances in Neural Information Processing Systems}, 35:\penalty0 1583--1595, 2022.

\bibitem[Goodfellow et~al.(2014)Goodfellow, Shlens, and Szegedy]{goodfellow2014explaining}
Ian~J Goodfellow, Jonathon Shlens, and Christian Szegedy.
\newblock Explaining and harnessing adversarial examples.
\newblock \emph{arXiv preprint arXiv:1412.6572}, 2014.

\bibitem[Grigorescu et~al.(2020)Grigorescu, Trasnea, Cocias, and Macesanu]{grigorescu2020survey}
Sorin Grigorescu, Bogdan Trasnea, Tiberiu Cocias, and Gigel Macesanu.
\newblock A survey of deep learning techniques for autonomous driving.
\newblock \emph{Journal of field robotics}, 37\penalty0 (3):\penalty0 362--386, 2020.

\bibitem[Guo et~al.(2017)Guo, Pleiss, Sun, and Weinberger]{guo2017calibration}
Chuan Guo, Geoff Pleiss, Yu Sun, and Kilian~Q Weinberger.
\newblock On calibration of modern neural networks.
\newblock In \emph{International Conference on Machine Learning}, pages 1321--1330. PMLR, 2017.

\bibitem[He et~al.(2016)He, Zhang, Ren, and Sun]{he2016deep}
Kaiming He, Xiangyu Zhang, Shaoqing Ren, and Jian Sun.
\newblock Deep residual learning for image recognition.
\newblock In \emph{Proceedings of the IEEE conference on Computer Vision and Pattern Recognition}, pages 770--778, 2016.

\bibitem[Jiang et~al.(2012)Jiang, Osl, Kim, and Ohno-Machado]{jiang2012calibrating}
Xiaoqian Jiang, Melanie Osl, Jihoon Kim, and Lucila Ohno-Machado.
\newblock Calibrating predictive model estimates to support personalized medicine.
\newblock \emph{Journal of the American Medical Informatics Association}, 19\penalty0 (2):\penalty0 263--274, 2012.

\bibitem[Joy et~al.(2023)Joy, Pinto, Lim, Torr, and Dokania]{joy2023sample}
Tom Joy, Francesco Pinto, Ser-Nam Lim, Philip~HS Torr, and Puneet~K Dokania.
\newblock Sample-dependent adaptive temperature scaling for improved calibration.
\newblock In \emph{Proceedings of the AAAI Conference on Artificial Intelligence}, pages 14919--14926, 2023.

\bibitem[Kang et~al.(2020)Kang, Jo, Nam, and Choi]{kang2020confidence}
Dongmin Kang, Yeonsik Jo, Yeongwoo Nam, and Jonghyun Choi.
\newblock Confidence calibration for incremental learning.
\newblock \emph{IEEE Access}, 8:\penalty0 126648--126660, 2020.

\bibitem[Kingma(2014)]{kingma2014adam}
Diederik~P Kingma.
\newblock Adam: A method for stochastic optimization.
\newblock \emph{arXiv preprint arXiv:1412.6980}, 2014.

\bibitem[Krizhevsky et~al.(2009)Krizhevsky, Hinton, et~al.]{krizhevsky2009learning}
Alex Krizhevsky, Geoffrey Hinton, et~al.
\newblock Learning multiple layers of features from tiny images.
\newblock 2009.

\bibitem[Kull et~al.(2019)Kull, Perello~Nieto, K{\"a}ngsepp, Silva~Filho, Song, and Flach]{kull2019beyond}
Meelis Kull, Miquel Perello~Nieto, Markus K{\"a}ngsepp, Telmo Silva~Filho, Hao Song, and Peter Flach.
\newblock Beyond temperature scaling: Obtaining well-calibrated multi-class probabilities with dirichlet calibration.
\newblock \emph{Advances in Neural Information Processing Systems}, 32, 2019.

\bibitem[Kuzborskij et~al.(2013)Kuzborskij, Orabona, and Caputo]{Kuzborskij_2013_CVPR}
Ilja Kuzborskij, Francesco Orabona, and Barbara Caputo.
\newblock From n to n+1: Multiclass transfer incremental learning.
\newblock In \emph{Proceedings of the IEEE Conference on Computer Vision and Pattern Recognition}, 2013.

\bibitem[Le and Yang(2015)]{le2015tiny}
Ya Le and Xuan Yang.
\newblock Tiny imagenet visual recognition challenge.
\newblock \emph{CS 231N}, 7\penalty0 (7):\penalty0 3, 2015.

\bibitem[Li et~al.(2024)Li, Piccoli, Cossu, Bacciu, and Lomonaco]{li2024calibration}
Lanpei Li, Elia Piccoli, Andrea Cossu, Davide Bacciu, and Vincenzo Lomonaco.
\newblock Calibration of continual learning models.
\newblock In \emph{Proceedings of the IEEE/CVF Conference on Computer Vision and Pattern Recognition Workshop (CLVISION)}, pages 4160--4169, 2024.

\bibitem[Liu et~al.(2022)Liu, Ben~Ayed, Galdran, and Dolz]{liu2022devil}
Bingyuan Liu, Ismail Ben~Ayed, Adrian Galdran, and Jose Dolz.
\newblock The devil is in the margin: Margin-based label smoothing for network calibration.
\newblock In \emph{Proceedings of the IEEE/CVF Conference on Computer Vision and Pattern Recognition}, pages 80--88, 2022.

\bibitem[Lopez-Paz and Ranzato(2017)]{lopez2017gradient}
David Lopez-Paz and Marc'Aurelio Ranzato.
\newblock Gradient episodic memory for continual learning.
\newblock \emph{Advances in Neural Information Processing Systems}, 30, 2017.

\bibitem[McCloskey and Cohen(1989)]{mccloskey1989catastrophic}
Michael McCloskey and Neal~J Cohen.
\newblock Catastrophic interference in connectionist networks: The sequential learning problem.
\newblock In \emph{Psychology of learning and motivation}, pages 109--165. Elsevier, 1989.

\bibitem[Mukhoti et~al.(2020)Mukhoti, Kulharia, Sanyal, Golodetz, Torr, and Dokania]{mukhoti2020calibrating}
Jishnu Mukhoti, Viveka Kulharia, Amartya Sanyal, Stuart Golodetz, Philip Torr, and Puneet Dokania.
\newblock Calibrating deep neural networks using focal loss.
\newblock \emph{Advances in Neural Information Processing Systems}, 33:\penalty0 15288--15299, 2020.

\bibitem[Naeini et~al.(2015)Naeini, Cooper, and Hauskrecht]{naeini2015obtaining}
Mahdi~Pakdaman Naeini, Gregory Cooper, and Milos Hauskrecht.
\newblock Obtaining well calibrated probabilities using bayesian binning.
\newblock In \emph{Proceedings of the AAAI conference on artificial intelligence}, 2015.

\bibitem[Noh et~al.(2023)Noh, Park, Lee, and Ham]{noh2023rankmixup}
Jongyoun Noh, Hyekang Park, Junghyup Lee, and Bumsub Ham.
\newblock Rankmixup: Ranking-based mixup training for network calibration.
\newblock In \emph{Proceedings of the IEEE/CVF International Conference on Computer Vision}, pages 1358--1368, 2023.

\bibitem[Ovadia et~al.(2019)Ovadia, Fertig, Ren, Nado, Sculley, Nowozin, Dillon, Lakshminarayanan, and Snoek]{ovadia2019can}
Yaniv Ovadia, Emily Fertig, Jie Ren, Zachary Nado, David Sculley, Sebastian Nowozin, Joshua Dillon, Balaji Lakshminarayanan, and Jasper Snoek.
\newblock Can you trust your model's uncertainty? evaluating predictive uncertainty under dataset shift.
\newblock \emph{Advances in Neural Information Processing Systems}, 32, 2019.

\bibitem[Park et~al.(2023)Park, Noh, Oh, Baek, and Ham]{park2023acls}
Hyekang Park, Jongyoun Noh, Youngmin Oh, Donghyeon Baek, and Bumsub Ham.
\newblock Acls: Adaptive and conditional label smoothing for network calibration.
\newblock In \emph{Proceedings of the IEEE/CVF International Conference on Computer Vision}, pages 3936--3945, 2023.

\bibitem[Paszke et~al.(2019)Paszke, Gross, Massa, Lerer, Bradbury, Chanan, Killeen, Lin, Gimelshein, Antiga, et~al.]{paszke2019pytorch}
Adam Paszke, Sam Gross, Francisco Massa, Adam Lerer, James Bradbury, Gregory Chanan, Trevor Killeen, Zeming Lin, Natalia Gimelshein, Luca Antiga, et~al.
\newblock Pytorch: An imperative style, high-performance deep learning library.
\newblock \emph{Advances in Neural Information Processing Systems}, 32, 2019.

\bibitem[Platt et~al.(1999)]{platt1999probabilistic}
John Platt et~al.
\newblock Probabilistic outputs for support vector machines and comparisons to regularized likelihood methods.
\newblock \emph{Advances in large margin classifiers}, 10\penalty0 (3):\penalty0 61--74, 1999.

\bibitem[Rebuffi et~al.(2017)Rebuffi, Kolesnikov, Sperl, and Lampert]{rebuffi2017icarl}
Sylvestre-Alvise Rebuffi, Alexander Kolesnikov, Georg Sperl, and Christoph~H Lampert.
\newblock icarl: Incremental classifier and representation learning.
\newblock In \emph{Proceedings of the IEEE conference on Computer Vision and Pattern Recognition}, pages 2001--2010, 2017.

\bibitem[Russakovsky et~al.(2015)Russakovsky, Deng, Su, Krause, Satheesh, Ma, Huang, Karpathy, Khosla, Bernstein, et~al.]{russakovsky2015imagenet}
Olga Russakovsky, Jia Deng, Hao Su, Jonathan Krause, Sanjeev Satheesh, Sean Ma, Zhiheng Huang, Andrej Karpathy, Aditya Khosla, Michael Bernstein, et~al.
\newblock Imagenet large scale visual recognition challenge.
\newblock \emph{International journal of computer vision}, 115:\penalty0 211--252, 2015.

\bibitem[Szegedy et~al.(2016)Szegedy, Vanhoucke, Ioffe, Shlens, and Wojna]{szegedy2016rethinking}
Christian Szegedy, Vincent Vanhoucke, Sergey Ioffe, Jon Shlens, and Zbigniew Wojna.
\newblock Rethinking the inception architecture for computer vision.
\newblock In \emph{Proceedings of the IEEE conference on Computer Vision and Pattern Recognition}, pages 2818--2826, 2016.

\bibitem[Tomani et~al.(2021)Tomani, Gruber, Erdem, Cremers, and Buettner]{tomani2021post}
Christian Tomani, Sebastian Gruber, Muhammed~Ebrar Erdem, Daniel Cremers, and Florian Buettner.
\newblock Post-hoc uncertainty calibration for domain drift scenarios.
\newblock In \emph{Proceedings of the IEEE/CVF Conference on Computer Vision and Pattern Recognition}, pages 10124--10132, 2021.

\bibitem[Wu et~al.(2019)Wu, Chen, Wang, Ye, Liu, Guo, and Fu]{wu2019large}
Yue Wu, Yinpeng Chen, Lijuan Wang, Yuancheng Ye, Zicheng Liu, Yandong Guo, and Yun Fu.
\newblock Large scale incremental learning.
\newblock In \emph{Proceedings of the IEEE/CVF conference on Computer Vision and Pattern Recognition}, pages 374--382, 2019.

\bibitem[Yan et~al.(2021)Yan, Xie, and He]{yan2021dynamically}
Shipeng Yan, Jiangwei Xie, and Xuming He.
\newblock Der: Dynamically expandable representation for class incremental learning.
\newblock In \emph{Proceedings of the IEEE/CVF conference on Computer Vision and Pattern Recognition}, pages 3014--3023, 2021.

\bibitem[Yu et~al.(2022)Yu, Bates, Ma, and Jordan]{yu2022robust}
Yaodong Yu, Stephen Bates, Yi Ma, and Michael Jordan.
\newblock Robust calibration with multi-domain temperature scaling.
\newblock \emph{Advances in Neural Information Processing Systems}, 35:\penalty0 27510--27523, 2022.

\bibitem[Zadrozny and Elkan(2001)]{zadrozny2001obtaining}
Bianca Zadrozny and Charles Elkan.
\newblock Obtaining calibrated probability estimates from decision trees and naive bayesian classifiers.
\newblock In \emph{International Conference on Machine Learning}, pages 609--616, 2001.

\bibitem[Zadrozny and Elkan(2002)]{zadrozny2002transforming}
Bianca Zadrozny and Charles Elkan.
\newblock Transforming classifier scores into accurate multiclass probability estimates.
\newblock In \emph{Proceedings of the eighth ACM SIGKDD international conference on Knowledge discovery and data mining}, pages 694--699, 2002.

\bibitem[Zhang et~al.(2018)Zhang, Ciss{\'{e}}, Dauphin, and Lopez{-}Paz]{DBLP:conf/iclr/ZhangCDL18}
Hongyi Zhang, Moustapha Ciss{\'{e}}, Yann~N. Dauphin, and David Lopez{-}Paz.
\newblock mixup: Beyond empirical risk minimization.
\newblock In \emph{International Conference on Learning Representations}, 2018.

\bibitem[Zhang et~al.(2020)Zhang, Kailkhura, and Han]{zhang2020mix}
Jize Zhang, Bhavya Kailkhura, and T Han.
\newblock Mix-n-match: Ensemble and compositional methods for uncertainty calibration in deep learning.
\newblock In \emph{International Conference on Machine Learning}, 2020.

\bibitem[Zhao et~al.(2020)Zhao, Xiao, Gan, Zhang, and Xia]{zhao2020maintaining}
Bowen Zhao, Xi Xiao, Guojun Gan, Bin Zhang, and Shu-Tao Xia.
\newblock Maintaining discrimination and fairness in class incremental learning.
\newblock In \emph{Proceedings of the IEEE/CVF conference on Computer Vision and Pattern Recognition}, pages 13208--13217, 2020.

\bibitem[Zhou et~al.(2023)Zhou, Wang, Qi, Ye, Zhan, and Liu]{zhou2023deep}
Da-Wei Zhou, Qi-Wei Wang, Zhi-Hong Qi, Han-Jia Ye, De-Chuan Zhan, and Ziwei Liu.
\newblock Deep class-incremental learning: A survey.
\newblock \emph{arXiv preprint arXiv:2302.03648}, 2023.

\bibitem[Zhou et~al.(2002)Zhou, Wu, and Tang]{zhou2002ensembling}
Zhi-Hua Zhou, Jianxin Wu, and Wei Tang.
\newblock Ensembling neural networks: many could be better than all.
\newblock \emph{Artificial intelligence}, 137\penalty0 (1-2):\penalty0 239--263, 2002.

\end{thebibliography}
}

\clearpage
\maketitlesupplementary
\setcounter{page}{1}
\setcounter{section}{0}

%%%%%%%%%%%%%%%%%%%%%%%%%%%%%%%%%%%%%%%%%%%%%%%%%%%%%%%%%%%%

\section{Detailed Algorithms}
We provide full algorithms of Algorithm~\ref{alg:binary_search} (\texttt{MagSearch}) for magnitude search, and Algorithm~\ref{alg:temp_opt} (\texttt{TempOpt}) for temperature optimization.

\begin{algorithm}[h]
\SetKwInput{Input}{Input}
\SetKwInOut{Output}{Output}
\SetNoFillComment
\Input{Model parameters $\theta = \{w, v\}$, exemplar set from new task $\mathcal{M}_{t, new}$, target temperature $T_{target}$, set of feature means $\mu$, tolerance $\delta$}
$\epsilon_{low} \leftarrow 0.0$, $\epsilon_{high} \leftarrow 1.0$ \\
\While{$\epsilon_{high} - \epsilon_{low} > \delta$}{
    $\epsilon = \frac{\epsilon_{low} + \epsilon_{high}}{2.0}$ \\
    $\mathcal{M}^{\epsilon}_{t, new} = \{\ \}$ \\
    \For {$(\mathbf{x}_e, y_e) \in \mathcal{M}_{t, new}$} {
        $y'_e = \displaystyle\argmax_{c : \mu_c \in \mu, c \neq y_e} \|\phi_v(\mathbf{x}_e) - \mu_c\|$ \\
        $\mathbf{x}^{adv}_e = \mathbf{x}_e - \epsilon \text{ sign}(\nabla_{\mathbf{x}_e}\mathcal{L}_{CE}(\mathbf{x}_e, y'_e ; \theta))$ \\
        $\mathcal{M}^{\epsilon}_{t, new} \leftarrow \mathcal{M}^{\epsilon}_{t, new} \cup \{(\mathbf{x}^{adv}_e, y_e)\}$
    }
    $T = \texttt{TempOpt}(\mathcal{M}^{\epsilon}_{t, new}, \theta)$ \\
    \eIf{$T < T_{target}$}{
        $\epsilon_{low} \leftarrow \epsilon$ \\
    }{
        $\epsilon_{high} \leftarrow \epsilon$ \\
    }
}
\Output{$\frac{\epsilon_{low} + \epsilon_{high}}{2.0}$}
\caption{The magnitude search algorithm for perturbation (\texttt{MagSearch}).}
\label{alg:binary_search}
\end{algorithm}

\begin{algorithm}[h]
\SetKwInput{Input}{Input}
\SetKwInOut{Output}{Output}
\SetNoFillComment
\Input{Dataset $\mathcal{D}$, model parameters $\theta$}
$T \leftarrow 1$ \tcp*{Initialize $T$}
\While{not converge}{
    \For{$(\mathbf{x}, y) \in \mathcal{D}$}{
        Update $T$ to minimize $\mathcal{L}_{CE}(\mathbf{x}, y ; \theta, T)$ \\
    }
}
\Output{$T$}
\caption{The temperature optimization algorithm (\texttt{TempOpt}).}
\label{alg:temp_opt}
\end{algorithm}

\section{Computational Complexity Analysis}

In this section, we analyze the computational complexity of \method{}. \method{} consists of four main components: temperature optimization, feature means calculation, perturbation magnitude search, and memory update.
The complexity of temperature optimization is $O(M)$, where $M$ represents the memory size, since we optimize the temperature on the set of perturbed exemplars for memory. Similarly, calculating feature means requires $O(M)$ operations. The adversarial search process takes $O(M)$ time (with a constant factor $k$ for perturbation iterations), and applying perturbations also has a complexity of $O(M)$. Therefore, the overall computational complexity of \method{} for a single incremental task is also $O(M)$. 

As conventional class-incremental learning approaches operate with $O(T(N_{\text{new}}+M))$ complexity, \method{} maintains a lighter $O(M)$ complexity, where $T$ is the number of tasks, $N_{\text{new}}$ is the number of new task data points, and $M$ is the memory size. Consequently, integrating \method{} with existing class-incremental frameworks preserves the asymptotic computational efficiency, as $M$ remains constant and substantially less than $N_{\text{new}}$. With a small overhead, \method{} is an efficient approach that can be practically combined with existing class-incremental learning methods.

\section{Experiments}

\subsection{Inapplicability of PerturbTS on the CIFAR-10}
The reason PerturbTS\,\citep{tomani2021post} is not applicable on the CIFAR-10 in a class-incremental learning setup is that, with only two new classes per task, the model quickly fits to the data, making it impossible to achieve the designated accuracy reduction through perturbation. This overfitting prevents the perturbation magnitude optimization from converging.

\subsection{Detailed Experimental Settings}
To obtain the best calibration performance of \method{} with the minimal impact on accuracy, we use a new-task validation set whose size varies depending on the class-incremental learning technique and dataset used. The new-task validation set sizes are listed in Table~\ref{tbl:val_size}. The effect of the new-task validation set size will be explained later.

Class-incremental learning techniques require specific training parameters. For both EEIL\,\citep{castro2018end} and WA\,\citep{zhao2020maintaining}, we set the knowledge distillation temperature to 2. EEIL and DER\,\citep{yan2021dynamically} incorporate a balanced fine-tuning phase. For this phase, we train the model for 30 epochs with 10 tasks and 100 epochs with 20 tasks.

After model training, we store a subset of new-task data used for training to the memory by uniformly sampling examples from each class. As the memory size is fixed, we remove some existing exemplars to accommodate the new-task data.

We make a fair comparison between post-hoc calibration methods including \method{} versus vanilla class-incremental learning techniques in terms of training data. In particular, whenever we take a (minimal) validation set from the training data, we only train models on the remaining training data. In comparison, the vanilla techniques always train on the full training set.

\begin{table}[t]
\caption{The size of the new-task validation set for each class-incremental learning method and dataset used in the main experiments.}
\centering
\begin{tabular}{lccc}
\toprule
Method & CIFAR-10 & CIFAR-100 & Tiny-ImageNet \\
\midrule
 ER & 500 & 500 & 100 \\
 EEIL & 300 & 300 & 200 \\
 WA & 100 & 500 & 100 \\
 DER & 500 & 1000 & 100 \\
\bottomrule
\end{tabular}
\label{tbl:val_size}
\end{table}

\subsection{Additional Experiments}
\paragraph{Full experimental results} We evaluate \method{} against five calibration baselines (Cal method) in combination with four class-incremental learning techniques (CIL method) across three datasets. Table~\ref{tbl:baselines_all} presents a comprehensive comparison of all possible combinations between post-hoc calibration methods and class-incremental learning techniques.

Overall, \method{} outperforms the five calibration baselines when integrated with four existing class-incremental learning techniques across three datasets. Notably, \method{} consistently shows low calibration errors compared to all the baselines. While PerturbTS achieves the best calibration performances when combined with EEIL and WA on the Tiny-ImageNet, its effectiveness is inconsistent. In addition, PerturbTS is not applicable to the CIFAR-10 dataset and exhibits unusually high calibration errors on the CIFAR-100 dataset. In contrast, \method{} demonstrates robust and superior performances across all experimental settings, consistently achieving lower calibration errors regardless of the underlying class-incremental learning technique. This comprehensive evaluation validates that \method{} is a more reliable and versatile approach for addressing calibration challenges in class-incremental learning scenarios.

\paragraph{Expansion of incremental tasks}
We evaluate the performance when expanding incremental tasks from 10 to 20 tasks, with results presented in Table~\ref{tbl:compatibility_20task}. For all class-incremental learning techniques, we use 100 samples from each new task as a validation set on both CIFAR-100 and Tiny-ImageNet.

The results show that \method{} outperforms most vanilla class-incremental learning techniques, with WA being the only exception. As explained in Section~\ref{sec:compatibility}, WA scales the output logits corresponding to the new task only after training. This scaling of specific logits may not align with the insight behind \method{}'s perturbation direction policy. Nevertheless, \method{} significantly improves the calibration performance of poorly calibrated class-incremental learning techniques.

\begin{figure*}[t]
\centering
\begin{subfigure}{\columnwidth}
\includegraphics[width=0.9\columnwidth]{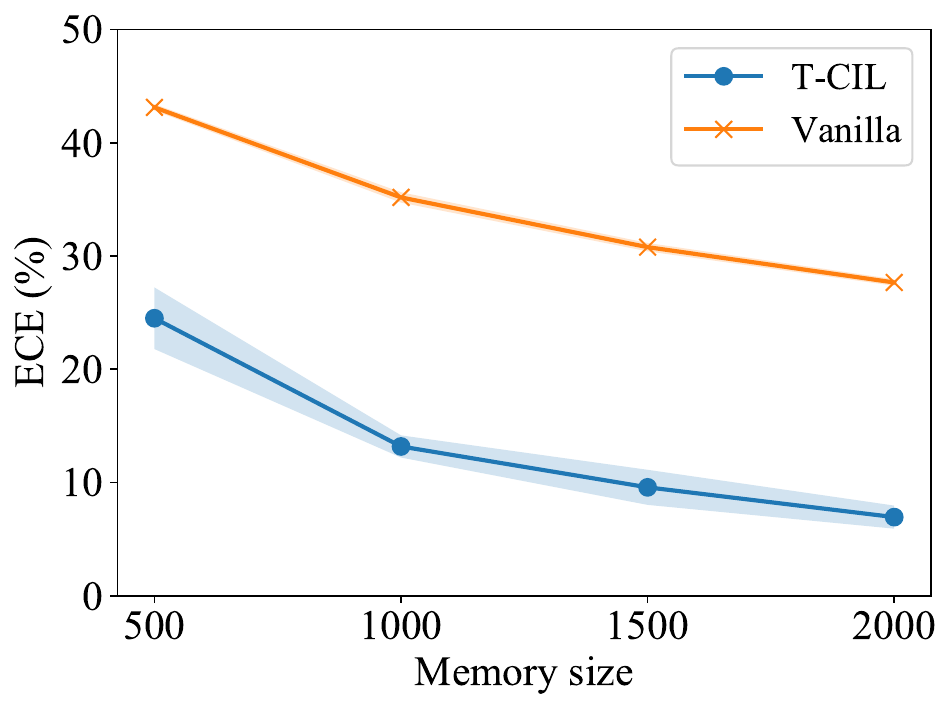}
\caption{}
\label{fig:mem_size_cifar}
\end{subfigure}
\begin{subfigure}{\columnwidth}
\includegraphics[width=0.9\columnwidth]{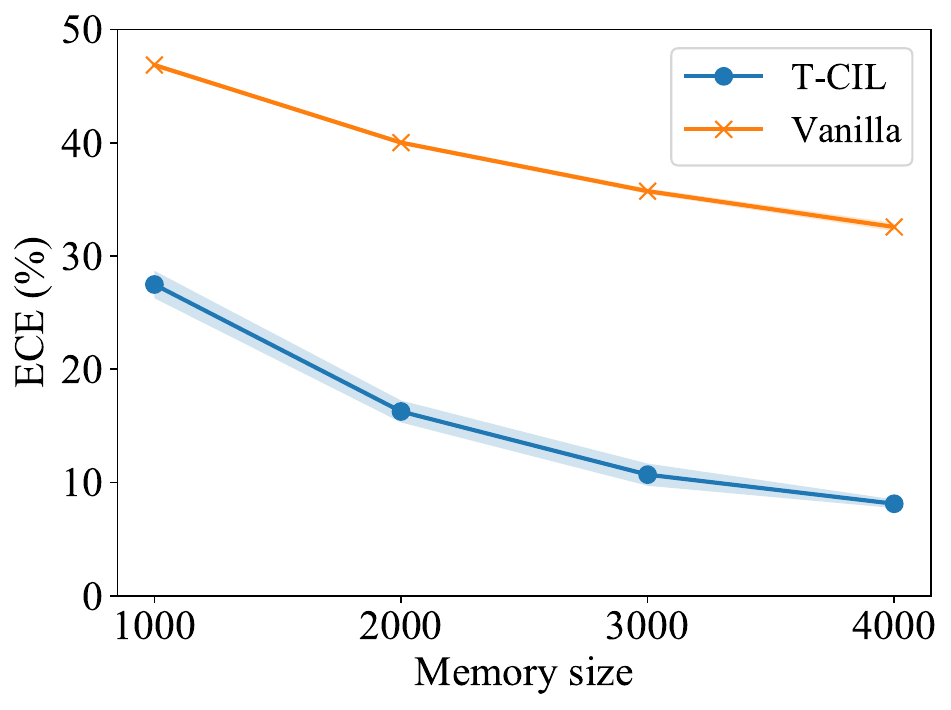}
\caption{}
\label{fig:mem_size_tiny}
\end{subfigure}
\caption{\method{} performance when varying the memory size on (a) CIFAR-100 and (b) Tiny-ImageNet.}
\label{fig:mem_size}
% \vskip -0.1in
\end{figure*}

\paragraph{Varying size of memory}
We analyze how memory size affects the ECE of \method{} compared to the vanilla method without calibration on the CIFAR-100 and Tiny-ImageNet, as shown in Figure~\ref{fig:mem_size}. Using ER as our base class-incremental learning technique, we experiment with memory sizes of 500, 1,000, 1,500, and 2,000 samples for CIFAR-100, and 1,000, 2,000, 3,000, and 4,000 samples for Tiny-ImageNet. We use a new-task validation set sized of 100.

Our results demonstrate that \method{} consistently achieves significantly lower ECE than the vanilla method across all memory sizes. Although smaller memory sizes lead to decreased model accuracy and consequently worse calibration performance, \method{} with just 500 memory samples still outperforms the vanilla method using 2,000 samples in terms of calibration quality.

\begin{figure}[t]
\centering
    \subfloat[]{
        {\includegraphics[scale=0.23]{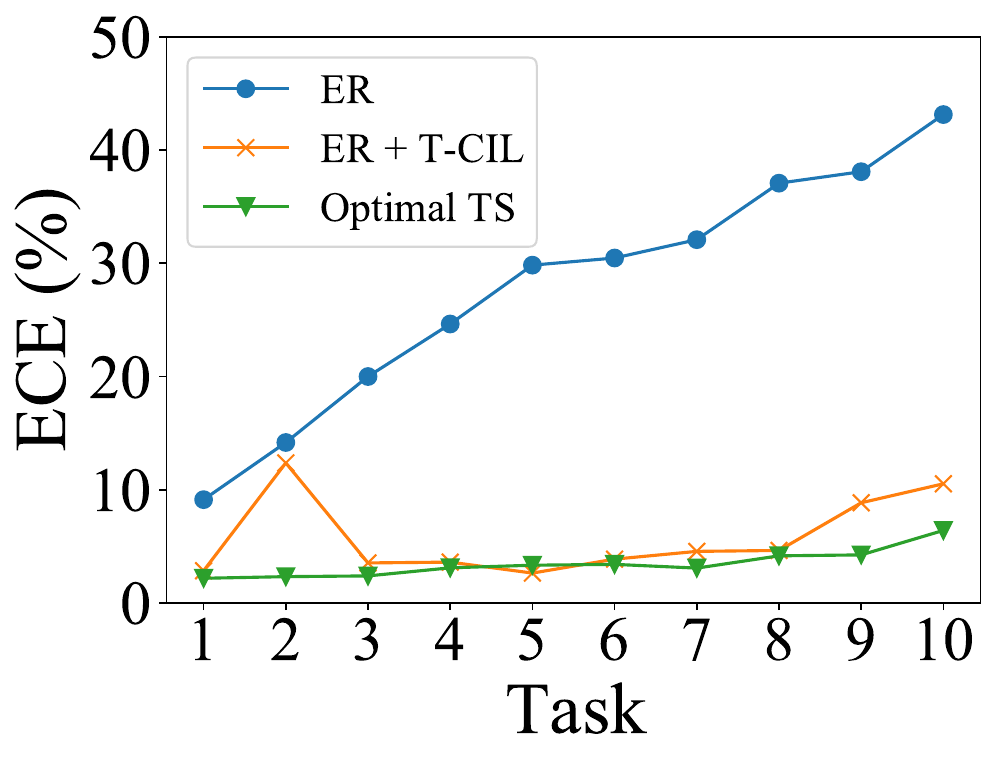}}
        \label{fig:ece_hist_ER}
        }
    % \hspace{0.1cm}
    \subfloat[]{
        {\includegraphics[scale=0.23]{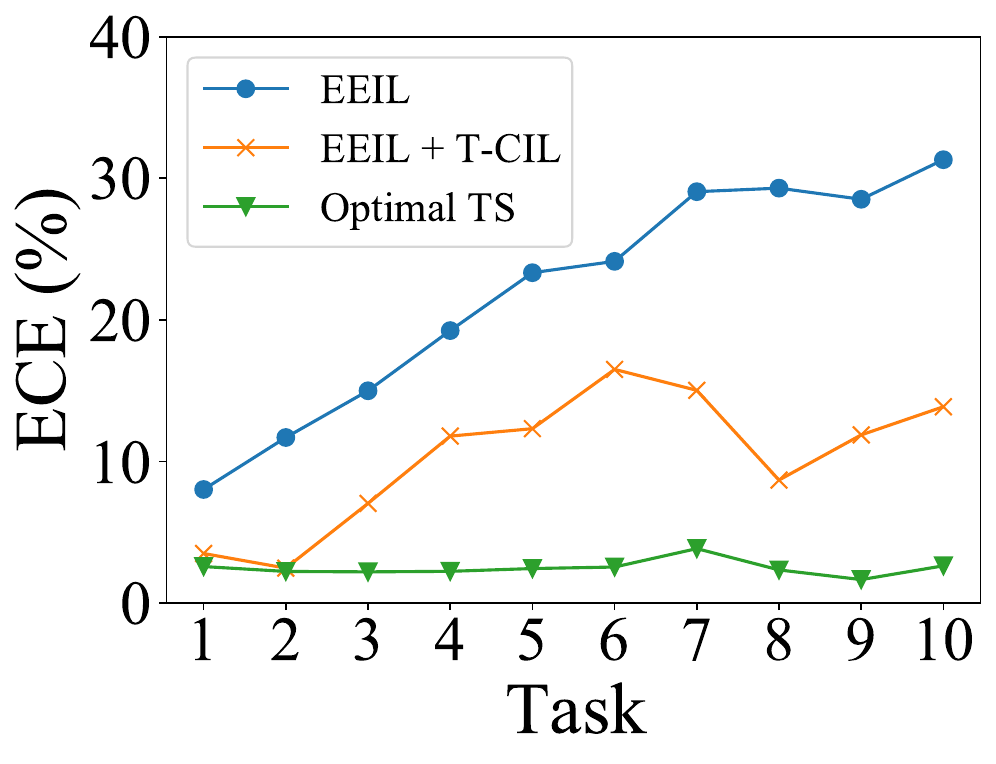}}
        \label{fig:ece_hist_EEIL}
        }   \\
    % \hspace{0.1cm}
    \subfloat[]{
        {\includegraphics[scale=0.23]{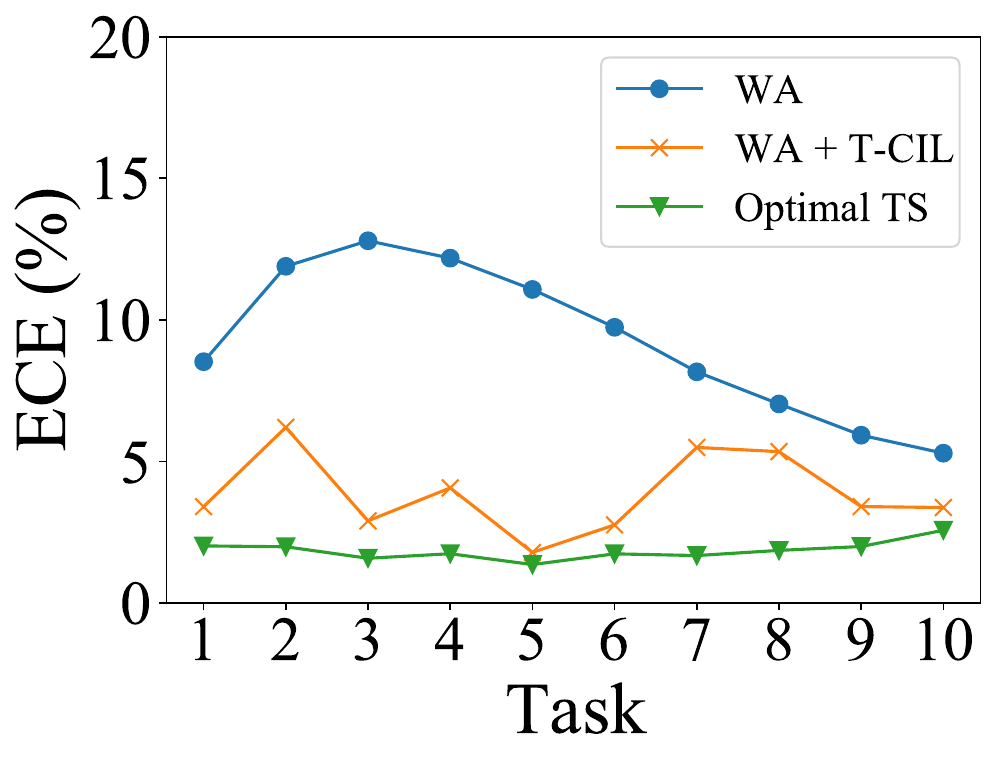}}
        \label{fig:ece_hist_WA}
        }
    % \hspace{0.1cm}
    \subfloat[]{
        {\includegraphics[scale=0.23]{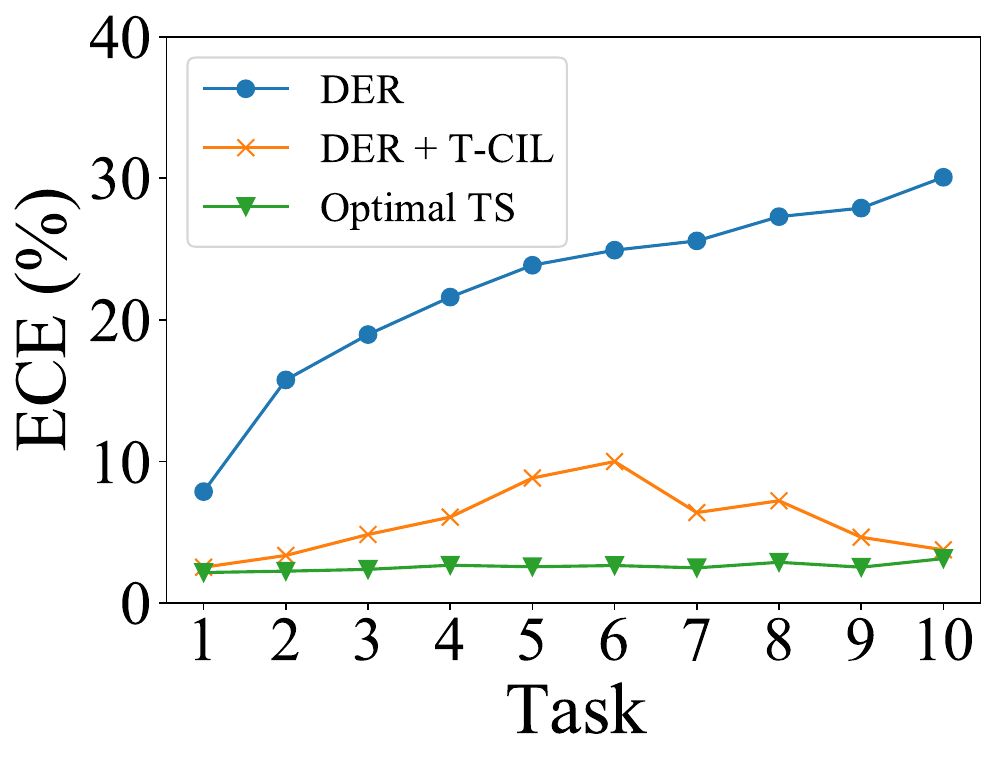}}
        \label{fig:ece_hist_DER}
        }
\vskip -0.1in
\caption{ECE (\%) comparison after training each task on the CIFAR-100 among existing class-incremental learning techniques, their combinations with \method{}, and the optimal TS. The existing techniques include: (a) ER, (b) EEIL, (c) WA, and (d) DER.}
\label{fig:ece_hist_cifar}
\vskip -0.1in
\end{figure}

\begin{figure*}[h]
\centering
\begin{minipage}[h]{0.8\columnwidth}
    \centering
    \includegraphics[width=\columnwidth]{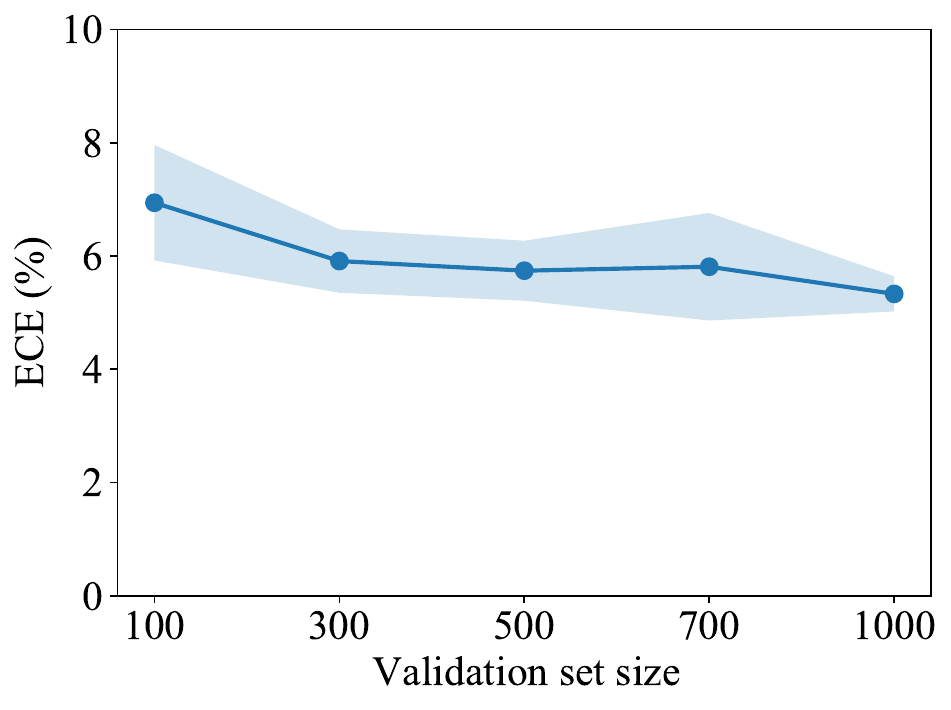}
    \caption{\method{} performance when varying the size of the new-task validation set on the CIFAR-100.}
    \label{fig:val_size}
\end{minipage}
\hfill
\begin{minipage}[h]{1.2\columnwidth}
    \centering
    \captionof{table}{\method{} performance when varying the size of the new-task validation set on the CIFAR-100.}
    \small
    \begin{tabular}{lccccc}
    \toprule
    Val size & 100 & 300 & 500 & 700 & 1000 \\
    \midrule
    ECE (\%) & {6.94}\tiny{$\pm$ 1.02} & {5.91}\tiny{$\pm$ 0.56} & {5.74}\tiny{$\pm$ 0.53} & {5.81}\tiny{$\pm$ 0.95} & {5.33}\tiny{$\pm$ 0.31} \\
    Acc (\%) & {56.97}\tiny{$\pm$ 0.65} & {56.95}\tiny{$\pm$ 0.34} & {56.25}\tiny{$\pm$ 0.62} & {56.55}\tiny{$\pm$ 0.80} & {56.08}\tiny{$\pm$ 0.57} \\
    \bottomrule
    \end{tabular}
    \label{tbl:varying_val_size}
\end{minipage}
\end{figure*}

\paragraph{Varying size of new-task validation set}
We vary the size of the new-task validation set and evaluate ECE and accuracy on CIFAR-100, using ER as a base class-incremental learning technique. We present the results in Figure~\ref{fig:val_size} and Table~\ref{tbl:varying_val_size}.

Figure~\ref{fig:val_size} demonstrates that \method{} is effective even with a small-sized validation set. As the validation set size increases, ECE decreases. 
However, increasing validation set size leads to accuracy drop due to the smaller training set size. These trends in ECE and accuracy indicate that \method{} only requires small new-task validation set for calibrating the model effectively while minimizing the impact on accuracy.

\begin{figure*}[t]
\centering
\begin{subfigure}{\columnwidth}
\includegraphics[width=\columnwidth]{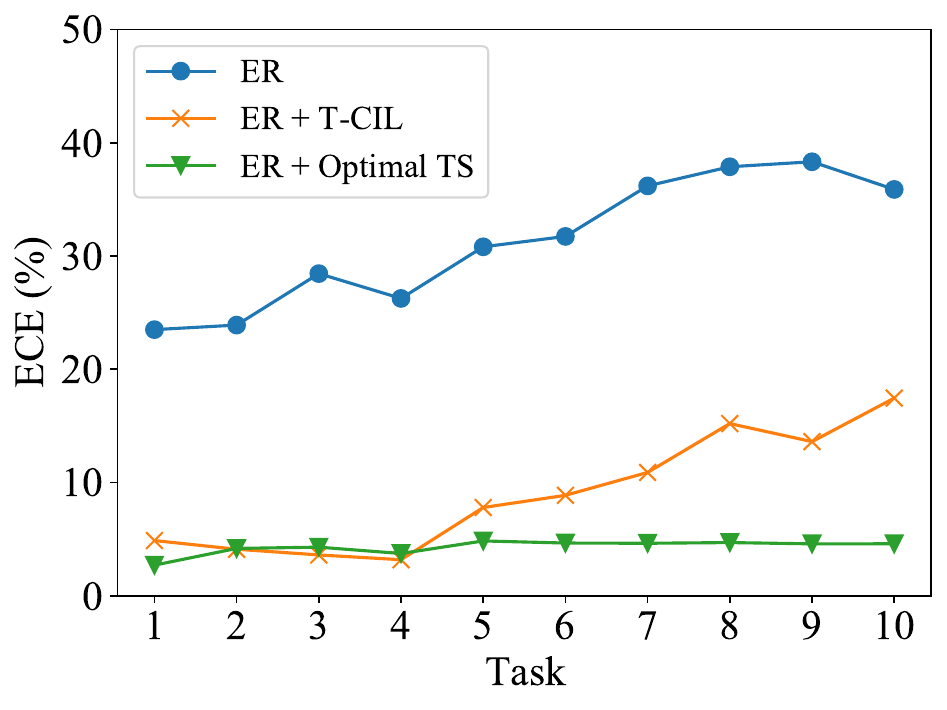}
\caption{}
\label{fig:acc_hist_er_tiny}
\end{subfigure}
\begin{subfigure}{\columnwidth}
\includegraphics[width=\columnwidth]{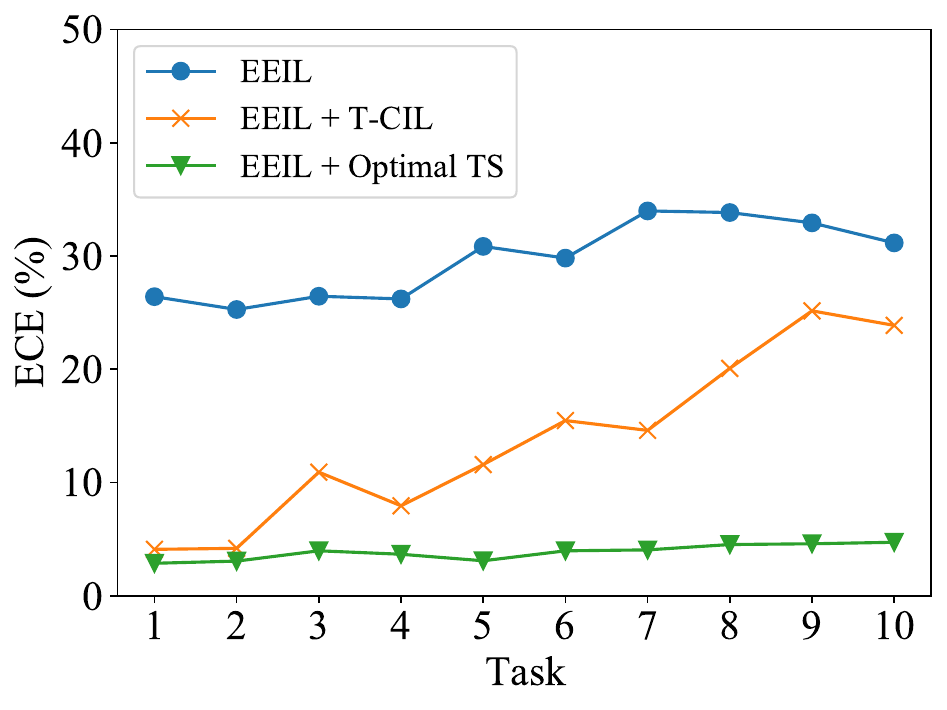}
\caption{}
\label{fig:ece_hist_eeil_tiny}
\end{subfigure}
\begin{subfigure}{\columnwidth}
\includegraphics[width=\columnwidth]{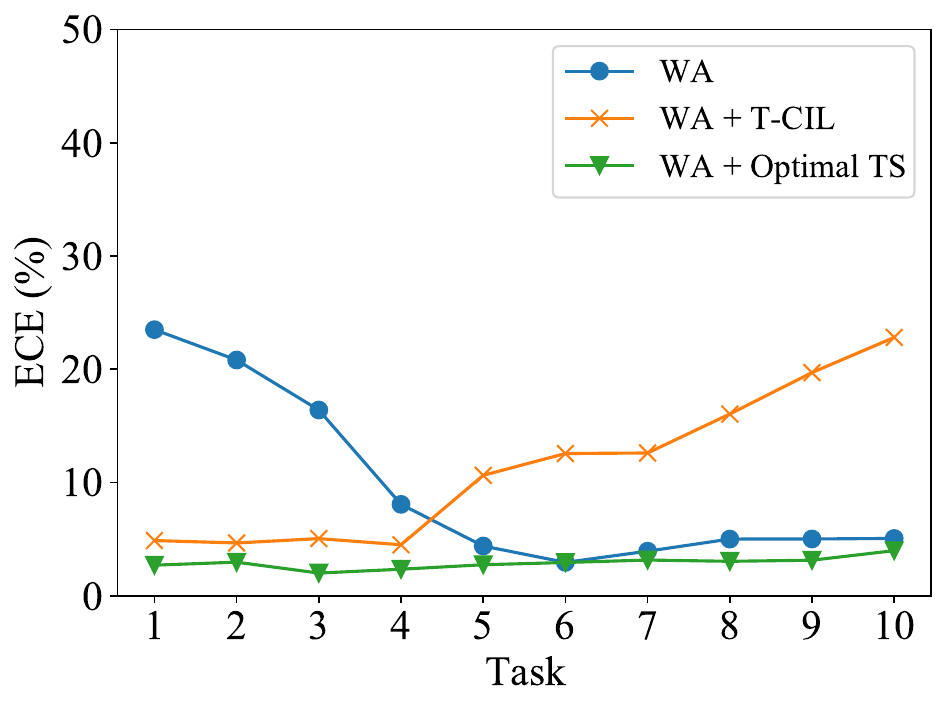}
\caption{}
\label{fig:acc_hist_wa_tiny}
\end{subfigure}
\begin{subfigure}{\columnwidth}
\includegraphics[width=\columnwidth]{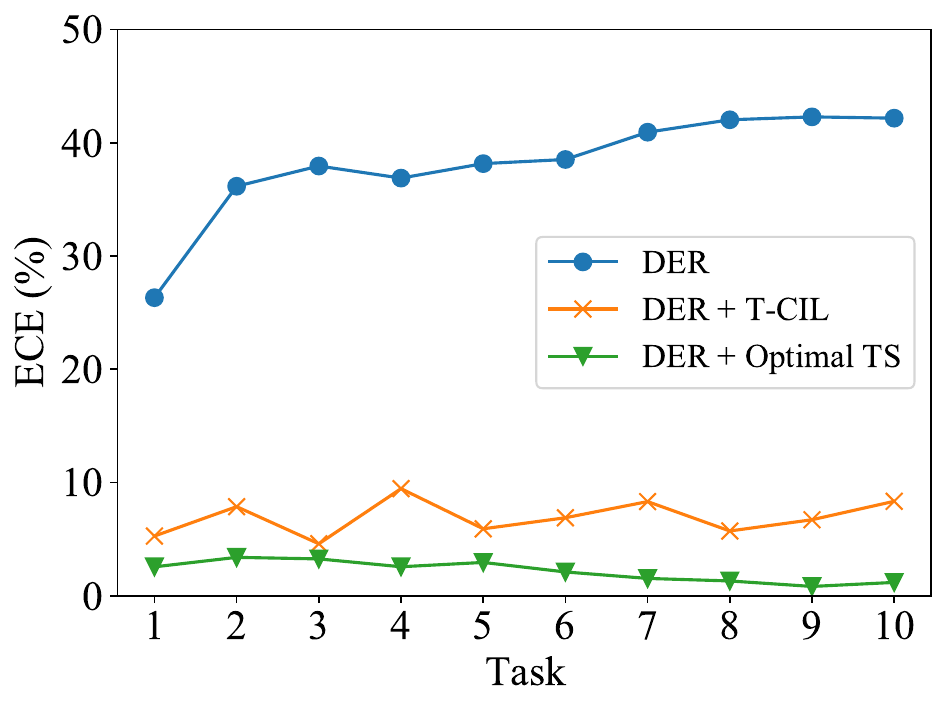}
\caption{}
\label{fig:acc_hist_der_tiny}
\end{subfigure}
\caption{ECE progression after training each task on the Tiny-ImageNet among existing class-incremental learning techniques, their combinations with \method{}, and the optimal TS. The existing techniques include: (a) ER, (b) EEIL, (c) WA, and (d) DER.}
\label{fig:ece_hist_tiny}
% \vskip -0.1in
\end{figure*}

\paragraph{Additional ECE progressions}
We present the progression of ECE across tasks on CIFAR-100 and Tiny-ImageNet of vanilla, \method{}, and Optimal TS when combined with four class-incremental learning techniques in Figure~\ref{fig:ece_hist_cifar} and Figure~\ref{fig:ece_hist_tiny}.
These figures show the progression of ECE throughout tasks, where we compare \method{} against an ideal scenario where we run \textit{TS} on the test set of both old and new tasks and thus obtain the best achievable calibration performance (called ``Optimal TS''). When combined with various class-incremental learning techniques, \method{} consistently demonstrates strong calibration performance across all tasks, with calibration errors approaching Optimal TS's ideal performance.

\begin{table*}[h]
\centering
\caption{Performance comparison between \method{} and five baselines when integrated with four class-incremental learning techniques on three datasets.}
\setlength{\tabcolsep}{3.4pt}
\vskip -0.05in
\small
\begin{tabular}{c|l|ccc|ccc|ccc}
\toprule
{} & {} & \multicolumn{3}{c|}{CIFAR-10} & \multicolumn{3}{c|}{CIFAR-100} & \multicolumn{3}{c}{Tiny-ImageNet} \\
\cmidrule{1-11}
{CIL Method} & {Cal Method} & {Acc. ($\uparrow$)} & {ECE ($\downarrow$)} & {AECE ($\downarrow$)} & {Acc. ($\uparrow$)} & {ECE ($\downarrow$)} & {AECE ($\downarrow$)} & {Acc. ($\uparrow$)} & {ECE ($\downarrow$)} & {AECE ($\downarrow$)}  \\
% \cmidrule{1-11}
\midrule
\multirow{6}{*}{ER\,\citep{chaudhry2019tiny}} & {Vanilla} & {65.61}\tiny{$\pm$ 0.49} & {28.16}\tiny{$\pm$ 0.33} & {28.12}\tiny{$\pm$ 0.32} & {56.51}\tiny{$\pm$ 0.37} & {27.66}\tiny{$\pm$ 0.29} & {27.64}\tiny{$\pm$ 0.28} & {31.95}\tiny{$\pm$ 0.54} & {32.56}\tiny{$\pm$ 0.36} & {32.55}\tiny{$\pm$ 0.35} \\
\cmidrule{2-11}
& {TS\,\citep{guo2017calibration}} & {65.86}\tiny{$\pm$ 0.09} & {23.97}\tiny{$\pm$ 0.82} & {23.92}\tiny{$\pm$ 0.81} & {56.25}\tiny{$\pm$ 0.62} & {16.28}\tiny{$\pm$ 0.32} & {16.22}\tiny{$\pm$ 0.37} & {31.48}\tiny{$\pm$ 0.39} & {19.44}\tiny{$\pm$ 0.75} & {19.44}\tiny{$\pm$ 0.75} \\
& {ETS\,\citep{zhang2020mix}} & {65.86}\tiny{$\pm$ 0.09} & {22.46}\tiny{$\pm$ 1.20} & {22.39}\tiny{$\pm$ 1.21} & {56.25}\tiny{$\pm$ 0.62} & {16.83}\tiny{$\pm$ 0.47} & {16.80}\tiny{$\pm$ 0.49} & {31.48}\tiny{$\pm$ 0.39} & {19.81}\tiny{$\pm$ 0.65} & {19.83}\tiny{$\pm$ 0.65} \\
& {IRM\,\citep{zhang2020mix}} & {65.86}\tiny{$\pm$ 0.09} & {22.09}\tiny{$\pm$ 1.73} & {21.83}\tiny{$\pm$ 1.73} & {56.25}\tiny{$\pm$ 0.62} & {17.50}\tiny{$\pm$ 0.64} & {17.39}\tiny{$\pm$ 0.61} & {31.48}\tiny{$\pm$ 0.39} & {19.77}\tiny{$\pm$ 0.72} & {19.73}\tiny{$\pm$ 0.78} \\
& {PerturbTS\,\citep{tomani2021post}} & {n/a} & {n/a} & {n/a} & {56.25}\tiny{$\pm$ 0.62} & {16.49}\tiny{$\pm$ 2.19} & {16.49}\tiny{$\pm$ 2.18} & {31.48}\tiny{$\pm$ 0.39} & {10.60}\tiny{$\pm$ 0.60} & {10.58}\tiny{$\pm$ 0.61} \\
& {\textbf{\method{}}} & {65.86}\tiny{$\pm$ 0.09} & \textbf{{17.70}\tiny{$\pm$ 2.60}} & \textbf{{17.64}\tiny{$\pm$ 2.59}} & {56.25}\tiny{$\pm$ 0.62} & \textbf{{5.74}\tiny{$\pm$ 0.53}} & \textbf{{5.75}\tiny{$\pm$ 0.50}} & {31.48}\tiny{$\pm$ 0.39} & \textbf{{8.12}\tiny{$\pm$ 0.38}} & \textbf{{8.12}\tiny{$\pm$ 0.41}} \\
\midrule
\multirow{6}{*}{EEIL\,\citep{castro2018end}} & {Vanilla} & {77.67}\tiny{$\pm$ 0.74} & {15.48}\tiny{$\pm$ 0.75} & {15.45}\tiny{$\pm$ 0.74} & {60.61}\tiny{$\pm$ 0.33} & {21.96}\tiny{$\pm$ 0.29} & {21.94}\tiny{$\pm$ 0.28} & {37.44}\tiny{$\pm$ 0.85} & {29.69}\tiny{$\pm$ 0.36} & {29.68}\tiny{$\pm$ 0.36} \\
\cmidrule{2-11}
& {TS} & {76.91}\tiny{$\pm$ 1.27} & \textbf{{10.20}\tiny{$\pm$ 0.94}} & \textbf{{10.16}\tiny{$\pm$ 0.93}} & {60.74}\tiny{$\pm$ 0.37} & {13.25}\tiny{$\pm$ 0.60} & {13.14}\tiny{$\pm$ 0.55} & {37.16}\tiny{$\pm$ 0.91} & {13.16}\tiny{$\pm$ 0.76} & {13.15}\tiny{$\pm$ 0.73} \\
& {ETS} & {76.91}\tiny{$\pm$ 1.27} & {10.31}\tiny{$\pm$ 0.71} & {10.29}\tiny{$\pm$ 0.72} & {60.74}\tiny{$\pm$ 0.37} & {13.89}\tiny{$\pm$ 0.47} & {13.83}\tiny{$\pm$ 0.42} & {37.16}\tiny{$\pm$ 0.91} & {13.52}\tiny{$\pm$ 0.73} & {13.51}\tiny{$\pm$ 0.71} \\
& {IRM} & {76.91}\tiny{$\pm$ 1.27} & {10.51}\tiny{$\pm$ 0.59} & {10.32}\tiny{$\pm$ 0.53} & {60.74}\tiny{$\pm$ 0.37} & {15.20}\tiny{$\pm$ 0.48} & {15.09}\tiny{$\pm$ 0.51} & {37.16}\tiny{$\pm$ 0.91} & {14.66}\tiny{$\pm$ 0.76} & {14.69}\tiny{$\pm$ 0.76} \\
& {PerturbTS} & {n/a} & {n/a} & {n/a} & {60.74}\tiny{$\pm$ 0.37} & {40.34}\tiny{$\pm$ 3.84} & {40.34}\tiny{$\pm$ 3.84} & {37.16}\tiny{$\pm$ 0.91} & \textbf{{8.81}\tiny{$\pm$ 0.59}} & \textbf{{8.81}\tiny{$\pm$ 0.61}} \\
& {\textbf{\method{}}} & {76.91}\tiny{$\pm$ 1.27} & {10.49}\tiny{$\pm$ 2.34} & {10.46}\tiny{$\pm$ 2.32} & {60.74}\tiny{$\pm$ 0.37} & \textbf{{10.30}\tiny{$\pm$ 1.10}} & \textbf{{10.22}\tiny{$\pm$ 1.06}} & {37.16}\tiny{$\pm$ 0.91} & {15.58}\tiny{$\pm$ 0.76} & {15.56}\tiny{$\pm$ 0.76} \\
\midrule
\multirow{6}{*}{WA\,\citep{zhao2020maintaining}} & {Vanilla} & {73.06}\tiny{$\pm$ 0.57} & {19.10}\tiny{$\pm$ 0.50} & {19.07}\tiny{$\pm$ 0.51} & {64.34}\tiny{$\pm$ 0.40} & {8.89}\tiny{$\pm$ 0.64} & {8.86}\tiny{$\pm$ 0.63} & {39.66}\tiny{$\pm$ 0.88} & {10.97}\tiny{$\pm$ 0.41} & {10.96}\tiny{$\pm$ 0.43} \\
\cmidrule{2-11}
& {TS} & {72.75}\tiny{$\pm$ 0.47} & {18.22}\tiny{$\pm$ 0.69} & {18.19}\tiny{$\pm$ 0.69} & {64.02}\tiny{$\pm$ 0.06} & {5.93}\tiny{$\pm$ 0.24} & {5.88}\tiny{$\pm$ 0.28} & {38.59}\tiny{$\pm$ 0.44} & {13.24}\tiny{$\pm$ 1.08} & {13.28}\tiny{$\pm$ 1.09} \\
& {ETS} & {72.75}\tiny{$\pm$ 0.47} & {17.96}\tiny{$\pm$ 0.67} & {17.92}\tiny{$\pm$ 0.69} & {64.02}\tiny{$\pm$ 0.06} & {5.77}\tiny{$\pm$ 0.39} & {5.75}\tiny{$\pm$ 0.42} & {38.59}\tiny{$\pm$ 0.44} & {13.01}\tiny{$\pm$ 1.12} & {13.06}\tiny{$\pm$ 1.12} \\
& {IRM} & {72.75}\tiny{$\pm$ 0.47} & {18.03}\tiny{$\pm$ 0.94} & {17.58}\tiny{$\pm$ 0.97} & {64.02}\tiny{$\pm$ 0.06} & {6.61}\tiny{$\pm$ 0.47} & {6.51}\tiny{$\pm$ 0.46} & {38.59}\tiny{$\pm$ 0.44} & {10.88}\tiny{$\pm$ 0.46} & {10.87}\tiny{$\pm$ 0.48} \\
& {PerturbTS} & {n/a} & {n/a} & {n/a} & {64.02}\tiny{$\pm$ 0.06} & {46.55}\tiny{$\pm$ 2.20} & {46.54}\tiny{$\pm$ 2.20} & {38.59}\tiny{$\pm$ 0.44} & \textbf{{8.63}\tiny{$\pm$ 1.07}} & \textbf{{8.61}\tiny{$\pm$ 1.10}} \\
& {\textbf{\method{}}} & {72.75}\tiny{$\pm$ 0.47} & \textbf{{15.61}\tiny{$\pm$ 0.23}} & \textbf{{15.58}\tiny{$\pm$ 0.22}} & {64.02}\tiny{$\pm$ 0.06} & \textbf{{3.87}\tiny{$\pm$ 0.52}} & \textbf{{3.84}\tiny{$\pm$ 0.55}} & {38.59}\tiny{$\pm$ 0.44} & {11.43}\tiny{$\pm$ 1.00} & {11.46}\tiny{$\pm$ 1.04} \\
\midrule
\multirow{6}{*}{DER\,\citep{yan2021dynamically}} & {Vanilla} & {74.53}\tiny{$\pm$  0.48} & {21.81}\tiny{$\pm$ 0.46} & {21.78}\tiny{$\pm$ 0.47} & {69.98}\tiny{$\pm$ 0.69} & {22.38}\tiny{$\pm$ 0.37} & {22.35}\tiny{$\pm$ 0.35} & {46.62}\tiny{$\pm$ 2.84} & {39.00}\tiny{$\pm$ 1.72} & {38.99}\tiny{$\pm$ 1.72} \\
\cmidrule{2-11}
& {TS} & {74.93}\tiny{$\pm$ 0.35} & {17.27}\tiny{$\pm$ 0.37} & {17.25}\tiny{$\pm$ 0.37} & {69.98}\tiny{$\pm$ 0.58} & {6.16}\tiny{$\pm$ 0.25} & {6.04}\tiny{$\pm$ 0.26} & {47.79}\tiny{$\pm$ 0.47} & {11.29}\tiny{$\pm$ 0.66} & {11.26}\tiny{$\pm$ 0.66} \\
& {ETS} & {74.93}\tiny{$\pm$ 0.35} & {16.82}\tiny{$\pm$ 0.28} & {16.79}\tiny{$\pm$ 0.29} & {69.98}\tiny{$\pm$ 0.58} & {6.12}\tiny{$\pm$ 0.36} & {6.03}\tiny{$\pm$ 0.37} & {47.79}\tiny{$\pm$ 0.47} & {7.83}\tiny{$\pm$ 0.63} & {7.86}\tiny{$\pm$ 0.58} \\
& {IRM} & {74.93}\tiny{$\pm$ 0.35} & {17.04}\tiny{$\pm$ 0.45} & {16.80}\tiny{$\pm$ 0.48} & {69.98}\tiny{$\pm$ 0.58} & {7.88}\tiny{$\pm$ 0.33} & {8.03}\tiny{$\pm$ 0.40} & {47.79}\tiny{$\pm$ 0.47} & {10.47}\tiny{$\pm$ 0.61} & {11.01}\tiny{$\pm$ 0.61} \\
& {PerturbTS} & {n/a} & {n/a} & {n/a} & {69.98}\tiny{$\pm$ 0.58} & {52.21}\tiny{$\pm$ 5.97} & {52.21}\tiny{$\pm$ 5.97} & {47.79}\tiny{$\pm$ 0.47} & {15.56}\tiny{$\pm$ 1.18} & {15.55}\tiny{$\pm$ 1.18} \\
& {\textbf{\method{}}} & {74.93}\tiny{$\pm$ 0.35} & \textbf{{12.70}\tiny{$\pm$ 1.35}} & \textbf{{12.68}\tiny{$\pm$ 1.34}} & {69.98}\tiny{$\pm$ 0.58} & \textbf{{4.37}\tiny{$\pm$ 0.59}} & \textbf{{4.34}\tiny{$\pm$ 0.60}} & {47.79}\tiny{$\pm$ 0.47} & \textbf{{6.91}\tiny{$\pm$ 0.86}} & \textbf{{6.90}\tiny{$\pm$ 0.84}} \\
\bottomrule
\end{tabular}
\label{tbl:baselines_all}
\end{table*}

\begin{table*}[h]
\centering
\caption{\method{} performance combined with four existing class-incremental learning techniques on two datasets, each containing 20 incremental tasks.}
\vskip -0.05in
% \small
\begin{tabular}{l|ccc|ccc}
\toprule
{} & \multicolumn{3}{c|}{CIFAR-100} & \multicolumn{3}{c}{Tiny-ImageNet} \\
\cmidrule{1-7}
{Method} & {Acc. ($\uparrow$)} & {ECE ($\downarrow$)} & {AECE ($\downarrow$)} & {Acc. ($\uparrow$)} & {ECE ($\downarrow$)} & {AECE ($\downarrow$)} \\
\midrule
{ER} & {54.81}\tiny{$\pm$ 0.70} & {28.95}\tiny{$\pm$ 0.84} & {28.93}\tiny{$\pm$ 0.84} & {30.43}\tiny{$\pm$ 0.51} & {34.90}\tiny{$\pm$ 0.16} & {34.88}\tiny{$\pm$ 0.16} \\
{ER+\method{}} & {54.48}\tiny{$\pm$ 1.95} & \textbf{{7.08}\tiny{$\pm$ 0.49}} & \textbf{{7.07}\tiny{$\pm$ 0.46}} & {30.62}\tiny{$\pm$ 0.29} & \textbf{{10.72}\tiny{$\pm$ 0.88}} & \textbf{{10.79}\tiny{$\pm$ 0.86}} \\
\cmidrule{1-7}
{EEIL} & {55.36}\tiny{$\pm$ 1.19} & {25.17}\tiny{$\pm$ 0.84} & {25.15}\tiny{$\pm$ 0.84} & {33.54}\tiny{$\pm$ 0.60} & {28.89}\tiny{$\pm$ 0.27} & {28.89}\tiny{$\pm$ 0.27} \\
{EEIL+\method{}} & {55.86}\tiny{$\pm$ 0.93} & \textbf{{14.38}\tiny{$\pm$1.34}} & \textbf{{14.35}\tiny{$\pm$ 1.35}} & {34.22}\tiny{$\pm$ 0.18} & \textbf{{24.45}\tiny{$\pm$ 0.67}} & \textbf{{24.46}\tiny{$\pm$ 0.67}} \\
\cmidrule{1-7}
{WA} & {59.30}\tiny{$\pm$ 0.64} & \textbf{{7.39}\tiny{$\pm$ 0.35}} & \textbf{{7.37}\tiny{$\pm$ 0.35}} & {35.99}\tiny{$\pm$ 0.97} & \textbf{{10.06}\tiny{$\pm$ 0.95}} & \textbf{{10.06}\tiny{$\pm$ 0.99}} \\
{WA+\method{}} & {58.89}\tiny{$\pm$ 0.67} & \textbf{{7.45}\tiny{$\pm$ 0.72}} & \textbf{{7.47}\tiny{$\pm$ 0.72}} & {36.80}\tiny{$\pm$ 0.42} & {19.56}\tiny{$\pm$ 0.47} & {19.57}\tiny{$\pm$ 0.48} \\
\cmidrule{1-7}
{DER} & {68.90}\tiny{$\pm$ 0.31} & {24.05}\tiny{$\pm$ 0.45} & {24.02}\tiny{$\pm$ 0.45} & {48.63}\tiny{$\pm$ 0.75} & {42.44}\tiny{$\pm$ 0.56} & {42.43}\tiny{$\pm$ 0.56} \\
{DER+\method{}} & {68.84}\tiny{$\pm$ 0.61} & \textbf{{5.78}\tiny{$\pm$ 0.92}} & \textbf{{5.75}\tiny{$\pm$ 0.91}} & {48.33}\tiny{$\pm$ 1.31} & \textbf{{5.90}\tiny{$\pm$ 0.97}} & \textbf{{5.91}\tiny{$\pm$ 0.96}} \\
\bottomrule
\end{tabular}
\label{tbl:compatibility_20task}
\vskip -0.1in
\end{table*}

\end{document}